\documentclass[pdflatex,sn-mathphys-num]{sn-jnl}

\usepackage{graphicx}%
\usepackage{multirow}%
\usepackage{amsmath,amssymb,amsfonts}%
\usepackage{amsthm}%
\usepackage{mathrsfs}%
\usepackage[title]{appendix}%
\usepackage{xcolor}%
\usepackage{textcomp}%
\usepackage{manyfoot}%
\usepackage{booktabs}%
\usepackage{algorithm}%
\usepackage{algorithmicx}%
\usepackage{algpseudocode}%
\usepackage{listings}%

\usepackage{caption}
\usepackage{algpseudocode}
\usepackage{tikz}
\usepackage{tikz-3dplot}
\usepackage{arydshln} 

\theoremstyle{thmstyleone}%
\newtheorem{theorem}{Theorem}
\newtheorem{proposition}[theorem]{Proposition}%

\theoremstyle{thmstyletwo}%

\theoremstyle{thmstylethree}%
\newtheorem{definition}{Definition}%

\begin{document}

\title[Article Title]{Solving Cyclic Antibandwidth Problem by SAT}


\author[1]{\fnm{Hieu} \sur{Truong Xuan}}\email{truongxuanhieu11@vnu.edu.vn}

\author*[1]{\fnm{Khanh} \sur{To Van}}\email{khanhtv@vnu.edu.vn}

\affil{\orgdiv{Faculty of Information Technology, \par University of Engineering and Technology}, \par \orgname{Vietnam National University}, \city{Hanoi}, \country{Vietnam}}


\abstract{
The Cyclic Antibandwidth Problem (CABP), a variant of the Antibandwidth Problem, is an NP-hard graph labeling problem with numerous applications. Despite significant research efforts, existing state-of-the-art approaches for CABP are exclusively heuristic or metaheuristic in nature, and exact methods have been limited to restricted graph classes.
In this paper, we present the first exact approach for the CABP on general graphs, based on SAT solving, called SAT-CAB. The proposed SAT-CAB method is able to systematically explore the solution space and guarantee global optimality, overcoming the limitations of previously reported heuristic algorithms. The proposed approach relies on a novel and efficient SAT encoding of CABP, in which the problem is transformed into a sequence of At-Most-One constraints. In particular, we introduce a compact representation of the At-Most-One constraints inherent to CABP, which significantly reduces the size of the resulting formulas and enables modern SAT solvers to effectively explore the solution space and to certify global optimality.
Extensive computational experiments on standard benchmark instances show that the proposed method efficiently solves CABP instances of practical relevance, while identifying several previously unknown optimal solutions. Moreover, global optimal cyclic antibandwidth values are proven for a number of benchmark instances for the first time. Comparative results indicate that the proposed exact method, SAT-CAB, consistently matches or surpasses the best-known solutions obtained by state-of-the-art heuristic algorithms such as MS-GVNS, HABC-CAB, and MACAB, as well as strong commercial Constraint Programming and Mixed Integer Programming solvers like CPLEX and Gurobi, particularly on general graphs, while also providing optimality guarantees. These results advance the state of the art for CABP and provide a new baseline for exact and hybrid methods on general graphs.
}

\keywords{Cyclic Antibandwidth, Graph labeling, SAT encoding, Sequential counter encoding, Cardinality constraints, At-Most-One constraints}

\maketitle

\section{Introduction}\label{sec1}

Graph layout problem (GLP) is a class of combinatorial optimization problems that aims to find a vertex arrangement, typically in the form of a linear ordering, of a given graph that optimizes specific objective functions.  
These problems have received significant attention due to their broad applicability across various domains such as parallel computer network optimization, Very Large Scale Integration (VLSI) circuit design, information retrieval, numerical analysis, computational biology, graph theory, scheduling, and archaelogy \cite{diaz2002graphlayoutsurvey}.

Antibandwidth Problem (ABP) is a well-known NP-hard variant of the GLP that has been extensively investigated in the literature. 
Its objective is to assign distinct integer labels to the vertices of a given graph, such that the minimum difference between the labels of any two connected vertices is maximized \cite{leung1984bandwidthvarians}. Intuitively, this can be viewed as placing vertices on a straight line while maximizing the distance between connected vertices. The ABP has numerous applications in scheduling and resource allocation \cite{leung1984bandwidthvarians}, including radio frequency assignment \cite{hale1980frequency}, obnoxious facility location \cite{cappanera1999survey}, and map coloring \cite{hu2010visualizing}.

Cyclic Antibandwidth Problem (CABP), also known as the cycle-separation problem \cite{leung1984bandwidthvarians}, is a variant of the ABP. 
Instead of arranging the vertices along a straight line, CABP places them on a cycle, where the distance between two vertices is defined as the minimum cyclic distance on the circle. 
The goal of CABP is to find a labeling function that maximizes the minimum cyclic distance between any pair of connected vertices. 
This cyclic formulation is particularly relevant in systems with periodic or circular structures, such as sensor network time slot allocation, cyclic scheduling, or circular memory allocation. 
Despite its practical relevance and theoretical interest, state-of-the-art approaches for CABP on general graphs are exclusively heuristic or metaheuristic, including approaches such as MACAB \cite{bansal2011memetic}, HABC-CAB \cite{lozano2013hybrid}, and MS-GVNS \cite{cavero2022general}, which aim to produce high-quality solutions but do not provide optimality guarantees. Existing exact approaches are limited to very restricted graph classes, such as \textit{path}, \textit{cycle}, \textit{two-dimensional mesh}, and \textit{toroidal mesh}, where the problem structure can be exploited. To the best of our knowledge, no exact solution method has been proposed for CABP on general graphs, and consequently, global optimal cyclic antibandwidth values remain unknown for most benchmark instances.

This gap motivates the need for an exact approach capable of solving CABP on general graphs and certifying global optimality. In this context, SAT-based approaches offer a promising exact framework, as they have proven effective for solving challenging combinatorial optimization and graph labeling problems. 
In particular, SAT solving has been successfully applied to the Antibandwidth Problem through specialized At-Most-One constraint encodings, such as Duplex encoding \cite{fazekas2020duplex} and SCL encoding \cite{Truong2025}, and has also demonstrated strong performance on related graph labeling problems, including graph coloring problems \cite{glorian2019incremental, schidler2023sat}. 

Recent advances in Boolean Satisfiability (SAT) solving, together with the development of increasingly efficient SAT solvers, have enabled a wide range of NP-hard combinatorial optimization problems to be formulated and solved within a satisfiability framework \cite{vardi2015sat}. Modern SAT solvers are capable of handling large-scale formulas and have become a viable exact alternative to heuristic and metaheuristic approaches in several optimization and graph labeling problems. SAT-based methods benefit from powerful propagation and decision mechanisms that allow an effective exploration of large search spaces and the certification of optimality. However, a key challenge in SAT-based modeling lies in controlling the size of the resulting formulas, as naïve encodings may generate a prohibitive number of variables and clauses. To address this issue, substantial research has focused on the design of compact and problem-specific encodings ~\cite{vasconcellos2020abacus, haberlandt2023effective}. 
Motivated by these developments, this paper investigates the feasibility of solving the Cyclic Antibandwidth Problem using a tailored SAT-based formulation. In particular, we introduce the Cyclic Ladder constraint, a novel constraint composed of overlapping At-Most-One (AMO) constraints over cyclic sequences of consecutive variables. By exploiting the shared variables between adjacent AMO constraints, we propose an efficient SAT encoding that jointly encodes these constraints, resulting in a more compact formulation than encoding each AMO constraint independently.

In summary, the contributions of this paper are as follows:

\begin{itemize}
    \item \textit{We propose the first exact approach for the Cyclic Antibandwidth Problem on general graphs, based on SAT solving, capable of systematically exploring the solution space and certifying global optimality.}
    \item \textit{We introduce a novel and compact SAT encoding tailored to the structure of CABP, based on an efficient representation of overlapping At-Most-One constraints, which significantly reduces the number of variables and clauses compared to direct encodings.}
    \item \textit{Extensive computational results show that the proposed approach improves upon the state of the art by discovering several previously unknown optimal solutions and by being the first to prove global optimal cyclic antibandwidth values for many standard benchmark instances.}
\end{itemize}

The remainder of this paper is organized as follows. Section 2 formally defines the Cyclic Antibandwidth Problem and reviews the related literature. Section 3 presents an exact formulation of CABP and introduces the SAT-based solving framework. Section 4 details the proposed Cyclic Ladder constraint together with its compact SAT encoding. Section 5 describes the experimental setup, including benchmark instances, implementation details, and evaluation criteria. Section 6 reports and discusses the computational results, comparing the proposed approach with state-of-the-art heuristic, Constraint Programming, and Mixed Integer Programming methods. Finally, Section 7 concludes the paper with a discussion of contributions, limitations, and potential applications.

\section{Problem Definition and Related Work}\label{sec2}

\subsection{Cyclic Antibandwidth Problem}

Given a graph $G = (V, E)$, where $V$ and $E$ denote the sets of vertices and edges of $G$, respectively. A labeling function $f$ of $G$ is a bijection $V \to \{1, \ldots , |V|\}$ that assigns a unique label to each vertex in $V$. The cyclic distance $D_{c}$ between two vertices $i$ and $i'$ with $\{i, i'\} \in E$ is defined as follows:
\[D_{c}(i, i') = \min (|f(i) - f(i')|, |V| - |f(i) - f(i')|).\]

The cyclic antibandwidth $CAB_{f}(G)$ of graph $G$ under the labeling function $f$ is the minimal value among the $D_{c}$ values of all the edges:
\[CAB_{f}(G) = \min_{(i, i')\in E} (D_{c}(i, i')).\]

The Cyclic Antibandwidth Problem aims to maximize the cyclic antibandwidth of $G$ over all possible labeling functions:
\[CAB(G) = \max_{f}CAB_{f}(G).\]

\subsection{Related Work}

The Cyclic Antibandwidth Problem (CABP), also known as the cycle-separation problem \cite{leung1984bandwidthvarians}, was first introduced by Leung et al. \cite{leung1984bandwidthvarians} with the objective of arranging the vertices of an undirected graph on a circle such that every pair of connected vertices has a distance greater than a given value $k$.
Since then, a variety of studies have been published to develop both theoretical results and heuristic methods for this problem across many different graph classes.
Table~\ref{tab:CABP-related-work} provides an overview of existing studies, including their key contributions, the graph classes addressed, and the availability of optimality guarantees for the obtained Cyclic Antibanwidth (CAB) values.

\newlength{\contributionwidth}
\setlength{\contributionwidth}{5.25cm}

\newlength{\targetgraphwidth}
\setlength{\targetgraphwidth}{5.75cm}

\begin{table*}[ht!]
    \caption{Related work on Cyclic Antibandwidth Problem.}
    \label{tab:CABP-related-work}
    \centering
    \resizebox{\textwidth}{!}{
    \begin{tabular}{|p{3.75cm}|p{\contributionwidth}|p{\targetgraphwidth}|p{2cm}|}
    \hline
    \multicolumn{1}{|c|}{\multirow{2}{*}{\textbf{Work}}} & \multicolumn{1}{c|}{\multirow{2}{*}{\textbf{Main contribution}}} & \multicolumn{1}{c|}{\multirow{2}{*}{\textbf{Target graph}}} & \multicolumn{1}{c|}{\textbf{Proof of}} \\
     & \multicolumn{1}{c|}{} & \multicolumn{1}{c|}{} & \multicolumn{1}{c|}{\textbf{Optimality}} \\ \hline
   \multirow{2}{*}{Sýkora et al., 2005 \cite{sykora2005cyclic}} & Theory CAB value for the fundamental graphs & Path, cycle, 2D mesh, toroidal mesh & Exact value \rule{0pt}{20pt} \\ \cline{2-4} 
    & Theory estimated CAB value for the hypercube & Hypercube & Estimating \rule{0pt}{20pt} \\ \hline
     \multirow{2}{*}{Raspaud et al., 2009 \cite{raspaud2009antibandwidth}} & \multirow{2}{*}{\parbox{\contributionwidth}{Proving the correctness of the values proposed by Sýkora et al. \cite{sykora2005cyclic}}} & Path, cycle, 2D mesh, toroidal mesh & Exact value \rule{0pt}{20pt} \\ \cline{3-4} 
     &  & Hypercube & Estimating \rule{0pt}{20pt} \\ \hline
    \multirow{2}{*}{Bansal et al., 2011 \cite{bansal2011memetic}} & \multirow{2}{*}{\parbox{\contributionwidth}{Proposing MACAB algorithm to solve CABP}} & \multirow{2}{*}{\parbox{\targetgraphwidth}{General graphs, including both structured and non-structured graphs}} & \multirow{2}{*}{Unprovable} \rule{0pt}{20pt} \\
    &  &  & \rule{0pt}{20pt} \\ \hline
    \multirow{2}{*}{Lozano et al., 2013 \cite{lozano2013hybrid}} & \multirow{2}{*}{\parbox{\contributionwidth}{Proposing HABC-CAB algorithm to solve CABP}} & \multirow{2}{*}{\parbox{\targetgraphwidth}{General graphs, including both structured and non-structured graphs}} & \multirow{2}{*}{Unprovable} \rule{0pt}{20pt}\\
     &  &  &  \rule{0pt}{20pt} \\ \hline
    \multirow{2}{*}{Sundar, 2019 \cite{sundar2019hybrid}} & \multirow{2}{*}{\parbox{\contributionwidth}{Proposing HACO-CAB algorithm to solve CABP}} & \multirow{2}{*}{\parbox{\targetgraphwidth}{Harwell-Boeing graph}} & \multirow{2}{*}{Unprovable}  \rule{0pt}{20pt}\\
     &  &  & \rule{0pt}{20pt} \\ \hline
    \multirow{2}{*}{Cavero et al., 2022 \cite{cavero2022general}} & \multirow{2}{*}{\parbox{\contributionwidth}{Proposing MS-GVNS algorithm to solve CABP}} & \multirow{2}{*}{\parbox{\targetgraphwidth}{General graphs, including both structured and non-structured graphs}} & \multirow{2}{*}{Unprovable} \rule{0pt}{20pt}\\
     &  &  & \rule{0pt}{20pt} \\ \hline
    \end{tabular}
    }
\end{table*}

In particular, Sýkora et al.~\cite{sykora2005cyclic} proposed exact formulas for determining the CAB value of several fundamental classes of graphs, including \textit{path}, \textit{cycle}, \textit{two-dimensional mesh}, and \textit{toroidal mesh}. In addition, they derived general upper and lower bounds of the CAB value for general graphs based on their Antibandwidth value. Furthermore, they presented an asymptotic formula for estimating the CAB value of \textit{hypercube}, accurate up to the third-order term as the dimension of the \textit{hypercube} tends to infinity. These results were later proven to be correct by Raspaud et al.~\cite{raspaud2009antibandwidth}.

To address the CABP on general graphs, Bansal et al. introduced the MACAB algorithm~\cite{bansal2011memetic}, a method based on the memetic algorithm \cite{moscato1989memeticalgorithm} that combines the genetic algorithm \cite{forrest1996geneticalgorithm} with local search to refine the solutions. MACAB first employs a breadth-first search (BFS) to generate initial solutions. After that, it applies a greedy label assignment heuristic on the BFS level structure to obtain promising CAB values. This heuristic is further enhanced with a genetic algorithm that incorporates crossover and mutation operators, providing a balance between intensification and diversification in the search process.

Inspired by the effectiveness of the social behavior of honeybees \cite{karaboga2005idea, karaboga2008performance, ziarati2011performance}, Lozano et al. proposed a population-based metaheuristic algorithm called HABC-CAB \cite{lozano2013hybrid} to address the CABP. This approach mimics the intelligent foraging behavior of honeybee swarms, which includes food searching, food evaluating, and food scouting. By mimicking this behavior, HABC-CAB helps quickly discover and efficiently exploit good labeling functions. In addition, HABC-CAB integrates tabu search \cite{glover1986tabusearch, xu1997tabu} to strengthen local intensification and exploitation, thereby improving its effectiveness in obtaining high-quality solutions for CABP instances.

HACO-CAB \cite{sundar2019hybrid} is a hybrid approach that combines the Ant Colony Optimization (ACO) \cite{dorigo1999ant} algorithm with a local search strategy to solve the CABP. In this approach, the pheromone deposition and evaporation mechanisms of ants are employed to guide the construction of Breadth-First Search spanning trees (BFS-ST), thereby exploring potential level structures of the graph. Based on these trees, the algorithm applies two different labeling schemes to generate feasible solutions and further refines them using a local search operator. By integrating the global exploration capability of ACO with the local exploitation ability of local search, HACO-CAB achieves high-quality solutions with improved CAB values and demonstrates competitive performance compared to previous approaches such as MACAB and HABC-CAB.

With the aim of overcoming local optimality, MS-GVNS \cite{cavero2022general} adopts a multistart framework to solve the CABP. In this framework, each iteration begins with the construction of an initial solution using a greedy breadth-first search procedure. After that, MS-GVNS applies a  random destruction–reconstruction shaking mechanism to diversify the search and explore previously unvisited regions of the solution space. The resulting solutions are then improved using General Variable Neighborhood Search (GVNS) \cite{mladenovic2008general, hansen2017variable}, which utilizes Variable Neighborhood Descent \cite{mladenovic1997variable} to explore a group of neighborhoods deterministically, such that the obtained result is a local optimum. Through this combination of diversification, intensification, and structural pruning, MS-GVNS achieves state-of-the-art performance across a wide range of benchmark instances.

Nevertheless, existing studies still suffer from several limitations. Although the works of Sýkora et al.~\cite{sykora2005cyclic} and Raspaud et al.~\cite{raspaud2009antibandwidth} present analytical formulations for determining optimal CAB values, these results are restricted to a narrow range of graph classes, mainly simple structures such as \textit{path}, \textit{cycle}, and \textit{two-dimensional mesh}. In contrast, metaheuristic methods, including MACAB, HABC-CAB, HACO-CAB, and MS-GVNS, are applicable to general graphs but inherently lack the ability to certify the optimality of the solutions they generate. Furthermore, the performance of such metaheuristic approaches is often sensitive to parameter tuning, which may significantly impact their effectiveness when applied to previously unseen or structurally different graph classes.

\section{SAT Solving for Cyclic Antibandwidth Problem}\label{sec3}

\subsection{Integer Linear Programming Model for CABP}
\label{sec:ILP_CABP}

This section introduces an Integer Linear Programming (ILP) formulation for the CABP. Let $x_i^l$ be a binary decision variable that is $true$ (or $1$) if vertex $i$ is assigned the label $l$, and $false$ (or $0$) otherwise. Constraints~\eqref{eq:vertices} and~\eqref{eq:labels} are imposed to guarantee that each label is assigned to exactly one vertex and that each vertex receives exactly one label, respectively. Together, these constraints enforce a one-to-one correspondence between vertices and labels, ensuring a valid labeling \cite{duarte2011grasp}.

\begin{align}
&\bigwedge_{l = 1}^{|V|} \big(\sum_{i = 1}^{|V|}x_{i}^{l}\big) = 1 \tag{VERTICES} \label{eq:vertices} \\
&\bigwedge_{i = 1}^{|V|} \big(\sum_{l = 1}^{|V|}x_{i}^{l}\big) = 1 \tag{LABELS} \label{eq:labels}
\end{align}

Inspired by the iterative model of Markus Sinnl \cite{sinnl2021note} for the Antibandwidth problem, we introduce a formulation to address the question: ``Does there exist a labeling such that $CAB(G) \geq k$?''. This formulation combines the two constraints \eqref{eq:vertices} and \eqref{eq:labels}, along with constraint \eqref{eq:cyclic-k}, which specifically enforces that the cyclic distance $D_c(i, i')$ of any edge $\{i, i'\} \in E$ is greater than or equal to $k$ whenever $k$ is a feasible CAB value.

\begin{equation}
\bigwedge_{l=1}^{|V|} 
    \big(\sum_{l'=0}^{k-1} (x_i^t + x_{i'}^t ) \leq 1\big),
\quad 
\left\{
\begin{aligned}
& \forall \{i, i'\} \in E, \\ 
& t = \big((l + l' - 1) \bmod |V|\big) + 1
\end{aligned}
\right.
\tag{CYCLIC-k}\label{eq:cyclic-k}
\end{equation}

Figure~\ref{fig:cyclic-8n-3k} illustrates the application of the constraint~\eqref{eq:cyclic-k} on an edge $\{i, i'\}$ in a graph with $|V| = 8$ and $k = 3$.
When an arbitrary label is assigned to vertex $i$, this constraint restricts the set of labels that cannot be assigned to vertex $i'$, and vise versa, ensuring that the cyclic distance between the two vertices is not smaller than~3.
For example, if $x_i^6$ is \textit{true}, constraint~\eqref{eq:cyclic-k} ensures that $x_{i'}^4$, $x_{i'}^5$, $x_{i'}^6$, $x_{i'}^7$, and $x_{i'}^8$ must be \textit{false}.
This implies that vertex $i'$ can only be assigned the label 1, 2, or 3.

\begin{figure}[ht!]
    \centering
    \resizebox{!}{!}{
    \begin{tabular}{p{0.1cm}p{0.1cm}p{0.1cm}p{0.1cm}p{0.1cm}p{0.1cm}p{0.1cm}p{0.1cm}p{0.1cm}p{0.1cm}p{0.1cm}p{1.5cm}}
    $x_{i}^{1}$ & $+$ & $x_{i}^{2}$ & $+$ & $x_{i}^{3}$ & $+$ & $x_{i'}^{1}$ & $+$ & $x_{i'}^{2}$ & $+$ & $x_{i'}^{3}$ & $\leq 1 \wedge$ \rule{0pt}{15pt} \\
    $x_{i}^{2}$ & $+$ & $x_{i}^{3}$ & $+$ & $x_{i}^{4}$ & $+$ & $x_{i'}^{2}$ & $+$ & $x_{i'}^{3}$ & $+$ & $x_{i'}^{4}$ & $\leq 1 \wedge$ \rule{0pt}{15pt} \\
    $x_{i}^{3}$ & $+$ & $x_{i}^{4}$ & $+$ & $x_{i}^{5}$ & $+$ & $x_{i'}^{3}$ & $+$ & $x_{i'}^{4}$ & $+$ & $x_{i'}^{5}$ & $\leq 1 \wedge$ \rule{0pt}{15pt} \\
    $x_{i}^{4}$ & $+$ & $x_{i}^{5}$ & $+$ & $x_{i}^{6}$ & $+$ & $x_{i'}^{4}$ & $+$ & $x_{i'}^{5}$ & $+$ & $x_{i'}^{6}$ & $\leq 1 \wedge$ \rule{0pt}{15pt} \\
    $x_{i}^{5}$ & $+$ & $x_{i}^{6}$ & $+$ & $x_{i}^{7}$ & $+$ & $x_{i'}^{5}$ & $+$ & $x_{i'}^{6}$ & $+$ & $x_{i'}^{7}$ & $\leq 1 \wedge$ \rule{0pt}{15pt} \\
    $x_{i}^{6}$ & $+$ & $x_{i}^{7}$ & $+$ & $x_{i}^{8}$ & $+$ & $x_{i'}^{6}$ & $+$ & $x_{i'}^{7}$ & $+$ & $x_{i'}^{8}$ & $\leq 1 \wedge$ \rule{0pt}{15pt} \\
    $x_{i}^{7}$ & $+$ & $x_{i}^{8}$ & $+$ & $x_{i}^{1}$ & $+$ & $x_{i'}^{7}$ & $+$ & $x_{i'}^{8}$ & $+$ & $x_{i'}^{1}$ & $\leq 1 \wedge$ \rule{0pt}{15pt} \\
    $x_{i}^{8}$ & $+$ & $x_{i}^{1}$ & $+$ & $x_{i}^{2}$ & $+$ & $x_{i'}^{8}$ & $+$ & $x_{i'}^{1}$ & $+$ & $x_{i'}^{2}$ & $\leq 1$ \rule{0pt}{15pt}
    \end{tabular}
    }
    \caption{\small \eqref{eq:cyclic-k} constraint of edge \{$i$, $i'$\} with $|V| = 8$ and $k=3$.}
    \label{fig:cyclic-8n-3k}
\end{figure}

Table~\ref{tab:d_c_{ii'}} presents the cyclic distance values $D_c(i,i')$ between vertices $i$ and $i'$ when vertex $i$ is assigned the label~6.
These cyclic distance values show that, under the constraint $D_c(i,i') \geq 3$, vertex $i'$ can only receive one of the labels 1, 2, or 3, which is consistent with the restriction enforced by constraint~\eqref{eq:cyclic-k}, as illustrated in Figure~\ref{fig:cyclic-8n-3k}.

\begin{table*}[ht!]
    \caption{Cyclic distance of edge $\{i, i'\}$ with vertex $i$ is assigned label 6.}
    \label{tab:d_c_{ii'}}
    \centering
    \resizebox{0.75\textwidth}{!}{
    \begin{tabular}{|c|c|c|c|c|c|}
    \hline
    $|V|$ & $f(i)$ & $f(i')$ & $|f(i) - f(i')|$ & $|V| - |f(i) - f(i')|$ & $D_c(i, i')$ \\ \hline
    8 & 6 & 1 & 5 & 3 & 3 \\ \hline
    8 & 6 & 2 & 4 & 4 & 4 \\ \hline
    8 & 6 & 3 & 3 & 5 & 3 \\ \hline
    8 & 6 & 4 & 2 & 6 & 2 \\ \hline
    8 & 6 & 5 & 1 & 7 & 1 \\ \hline
    8 & 6 & 6 & 0 & 8 & 0 \\ \hline
    8 & 6 & 7 & 1 & 7 & 1 \\ \hline
    8 & 6 & 8 & 2 & 6 & 2 \\ \hline
    \end{tabular}
    }
\end{table*}

\subsection{Iterative SAT Solving for CABP}
\label{sec:iterative-sat-solving}

To exactly solve the Cyclic Antibandwidth Problem, we adopt an iterative SAT-based algorithm that incrementally examines candidate CAB values within a given bounded range $[LB, UB]$. The key idea is to transform the CABP decision procedure, which determines whether there exists a labeling such that $CAB(G)\geq k$, into a sequence of SAT instances, each corresponding to a specific value of k.

The iterative SAT-based algorithm works as described in Algorithm~\ref{alg:sat_cab_sequential}. Starting from a known lower bound \(LB\) and progressively increasing up to a given upper bound \(UB\), the algorithm iteratively considers each candidate value \(k'\). For each \(k'\), the corresponding ILP formulation (given in Section~\ref{sec:ILP_CABP}) is instantiated with \(k = k'\) and subsequently encoded into a Boolean satisfiability (SAT) formula \(\Phi_{k'}\). This formula is then submitted to a SAT solver to search for satisfiability. If \(\Phi_{k'}\) is satisfiable, a feasible labeling achieving a CAB value of at least \(k'\) exists, and \(k'\) is recorded as the feasible CAB value. Otherwise, no labeling can achieve a CAB value of \(k'\), and the search terminates, as all larger values of \(k'\) are also infeasible. Consequently, the algorithm explores candidate values sequentially, systematically covering all values from \(LB\) up to the first infeasible value.

\begin{algorithm}[ht!]
\caption{Iterative SAT solving for CABP}
\label{alg:sat_cab_sequential}
\begin{algorithmic}[1]
\Require 
Graph $G = \{V,E\}$, SAT solver $\mathcal{S}$, lower bound $LB$, upper bound $UB$
\Ensure 
Optimal CAB value $k_{opt}$ and optimal labeling

\State $k_{best} = LB$
\For{$k' \gets LB + 1$ \textbf{to} $UB$}
    \State Encode the ILP model with $k$ = $k'$ into a CNF formula $\Phi_{k'}$
    \State Run SAT solver $\mathcal{S}$ on $\Phi_{k'}$
    \If{$\Phi_{k'}$ is \textbf{SAT}}
        \State Extract a feasible labeling $l_{k'}$
        \State $k_{best} \gets k'$
    \Else
        \State \textbf{break} \hfill $\triangleright$ No feasible solution for larger $k'$
    \EndIf
\EndFor

\State \textbf{return} $k_{best}$ as $k_{opt}$ and corresponding labeling
\end{algorithmic}
\end{algorithm}

The correctness of this algorithm follows directly from the monotonicity property of the CABP decision problem. Specifically, if no feasible labeling exists for a given value \(k'\), then no feasible labeling can exist for any value \(k'' > k'\). Consequently, if $k_{opt}$ is a feasible CAB value while $k_{opt} + 1$ is infeasible, then $k_{opt}$ is the optimal CAB value. Therefore, assuming that each SAT instance is solved to completion (i.e., without timeout), the iterative SAT-solving algorithm is guaranteed to identify the optimal CAB value and to certify its optimality by providing an explicit UNSAT proof for $k_{opt} + 1$.

\subsection{Decomposition of \eqref{eq:cyclic-k} constraint}

Constraint~\eqref{eq:cyclic-k} can be decomposed into two sequences of At-Most-One (AMO) constraints (also referred to as the Cyclic Ladder constraint in Section~\ref{sec:sat_encoding_for_cyclic_ladder_constraint}), each associated with one of the vertices \(i\) and \(i'\). This decomposition employs Proposition~\ref{prop:decompose_AMO_constraint}, which shows that any AMO constraint can be split into two sub-AMO constraints, together with an additional OR connecting clause that preserves the correctness of the original constraint.

\begin{proposition}
    The constraint $x_{1} + x_{2} + \ldots + x_{w} \leq 1$ holds if and only if for all \(i\) such that \(1 \leq i < w\) :
    \[(x_{1} + \ldots + x_{i}\leq 1)\wedge (x_{i+1} + \ldots + x_{w}\leq 1)\]
    \[\wedge (x_{1} + \ldots + x_{i}\leq 0 \vee x_{i+1} + \ldots + x_{w}\leq 0)\]
    \label{prop:decompose_AMO_constraint}
\end{proposition}

\begin{figure}[ht]
    \centering
    \resizebox{\textwidth}{!}{
    \begin{tabular}{cccccccccccccccl}
    \multicolumn{7}{c}{\textit{Vertex $i$}} & \multicolumn{1}{l}{} & \multicolumn{7}{c}{\textit{Vertex $i'$}} &  \\ \cdashline{1-7} \cdashline{9-15}
    \multicolumn{1}{:c}{$x_{i}^{1}$} & + & $x_{i}^{2}$ & + & \ldots & + & \multicolumn{1}{c:}{$x_{i}^{w}$} & \multicolumn{1}{c:}{+} & $x_{i'}^{1}$ & + & $x_{i'}^{2}$ & + & \ldots & + & \multicolumn{1}{c:}{$x_{i'}^{w}$} & $\leq 1 \wedge$ \rule{0pt}{15pt} \\
    \multicolumn{1}{:c}{$x_{i}^{2}$} & + & $x_{i}^{3}$ & + & \ldots & + & \multicolumn{1}{c:}{$x_{i}^{w+1}$} & \multicolumn{1}{c:}{+} & $x_{i'}^{2}$ & + & $x_{i'}^{3}$ & + & \ldots & + & \multicolumn{1}{c:}{$x_{i'}^{w+1}$} & $\leq 1 \wedge$ \rule{0pt}{15pt} \\
    \multicolumn{1}{:c}{} &  &  &  &  &  & \multicolumn{1}{c:}{} & \multicolumn{1}{c:}{\ldots} &  &  &  &  &  &  & \multicolumn{1}{c:}{} & \rule{0pt}{15pt} \\
    \multicolumn{1}{:c}{$x_{i}^{n-1}$} & + & $x_{i}^{n}$ & + & \ldots & + & \multicolumn{1}{c:}{$x_{i}^{w-2}$} & \multicolumn{1}{c:}{+} & $x_{i'}^{n-1}$ & + & $x_{i'}^{n}$ & + & \ldots & + & \multicolumn{1}{c:}{$x_{i'}^{w-2}$} & $\leq 1 \wedge$ \rule{0pt}{15pt} \\
    \multicolumn{1}{:c}{$x_{i}^{n}$} & + & $x_{i}^{1}$ & + & \ldots & + & \multicolumn{1}{c:}{$x_{i}^{w-1}$} & \multicolumn{1}{c:}{+} & $x_{i'}^{n}$ & + & $x_{i'}^{1}$ & + & \ldots & + & \multicolumn{1}{c:}{$x_{i'}^{w-1}$} & $\leq 1$ \rule{0pt}{15pt} \\ \cdashline{1-7} \cdashline{9-15}
    \end{tabular}
    }
    \caption{\eqref{eq:cyclic-k} constraint of edge $\{i, i'\}$.}
    \label{fig:cyclic_n_w}
\end{figure}

\begin{figure}[ht]
    \centering
    \begin{tabular}{c}
     $x_{i}^{1} + x_{i}^{2} + \ldots + x_{i}^{w} + x_{i'}^{1} + x_{i'}^{2} + \ldots + x_{i'}^{w} \leq 1$ \rule{0pt}{15pt}\\
     $\equiv$ \rule{0pt}{15pt}\\
     $(x_{i}^{1} + x_{i}^{2} + \ldots + x_{i}^{w} \leq 1) \wedge (x_{i'}^{1} + x_{i'}^{2} + \ldots + x_{i'}^{w} \leq 1)$ \rule{0pt}{15pt}\\
     $\wedge \big((x_{i}^{1} + x_{i}^{2} + \ldots + x_{i}^{w} \leq 0) \vee (x_{i'}^{1} + x_{i'}^{2} + \ldots + x_{i'}^{w} \leq 0)\big)$ \rule{0pt}{15pt}
    \end{tabular}
    \caption{Decomposition of the first AMO constraint in \eqref{eq:cyclic-k} constraint.}
    \label{fig:decompose_amo_constraint_i_i'_w}
\end{figure}

\begin{figure}[ht!]
\centering

\begin{minipage}{0.49\textwidth}
    \resizebox{\textwidth}{!}{
    \begin{tabular}{cccccccc}
    $x_{i}^{1}$ & + & $x_{i}^{2}$ & + & \ldots & + & $x_{i}^{w}$ & $\leq 1 \wedge$ \\[4pt]
    $x_{i}^{2}$ & + & $x_{i}^{3}$ & + & \ldots & + & $x_{i}^{w+1}$ & $\leq 1 \wedge$ \\[4pt]
     &  &  & \ldots &  &  &  &  \\[4pt]
    $x_{i}^{n-1}$ & + & $x_{i}^{n}$ & + & \ldots & + & $x_{i}^{w-2}$ & $\leq 1 \wedge$ \\[4pt]
    $x_{i}^{n}$ & + & $x_{i}^{1}$ & + & \ldots & + & $x_{i}^{w-1}$ & $\leq 1$
    \end{tabular}
    }
    \caption{Sequence of AMO constraints associated with vertex $i$.}
    \label{fig:cyclic_ladder_i}
\end{minipage}
\hfill
\begin{minipage}{0.49\textwidth}
    \resizebox{\textwidth}{!}{
    \begin{tabular}{cccccccc}
    $x_{i'}^{1}$ & + & $x_{i'}^{2}$ & + & \ldots & + & $x_{i'}^{w}$ & $\leq 1 \wedge$ \\[4pt]
    $x_{i'}^{2}$ & + & $x_{i'}^{3}$ & + & \ldots & + & $x_{i'}^{w+1}$ & $\leq 1 \wedge$ \\[4pt]
     &  &  & \ldots &  &  &  &  \\[4pt]
    $x_{i'}^{n-1}$ & + & $x_{i'}^{n}$ & + & \ldots & + & $x_{i'}^{w-2}$ & $\leq 1 \wedge$ \\[4pt]
    $x_{i'}^{n}$ & + & $x_{i'}^{1}$ & + & \ldots & + & $x_{i'}^{w-1}$ & $\leq 1$
    \end{tabular}
    }
    \caption{Sequence of AMO constraints associated with vertex $i'$.}
    \label{fig:cyclic_ladder_i'}

\end{minipage}

\end{figure}

For instance, given a \eqref{eq:cyclic-k} constraint associated with an edge $\{i, i'\}$, where $|V| = n$ and $k = w$, as illustrated in Figure~\ref{fig:cyclic_n_w}.
The Proposition~\ref{prop:decompose_AMO_constraint} decomposes each AMO constraint within this \eqref{eq:cyclic-k} constraint into two sub-constraints that involve exclusively the variables of vertex $i$ or $i'$, together with an OR clause that preserves the AMO condition (see Figure~\ref{fig:decompose_amo_constraint_i_i'_w}).
Grouping these sub-constraints forms two sequences of AMO constraints, each corresponding to one of the vertices, as shown in Figures~\ref{fig:cyclic_ladder_i} and~\ref{fig:cyclic_ladder_i'}, and a set of OR connecting clauses.

\subsection{Symmetry Breaking for CABP}
\label{sec:symmetry-breaking}

As presented by Sinnl \cite{sinnl2021note}, for any labeling function $f$ of an $n$-vertex graph, there always exists a reversed labeling function $f'$ in which each label in $f'$ corresponds one-to-one with a label in $f$ through the following linear transformation:

\[f' = n + 1 - f.\]

This transformation ensures that if $f$ satisfies \eqref{eq:vertices} and \eqref{eq:labels} constraints, then $f'$ also satisfies them. For instance, when $n = 7$ and $f \mapsto \{1,2,3,4,5,6,7\}$, the reversed labeling function $f'$ is $f' \mapsto \{7,6,5,4,3,2,1\}$, which still fulfills both \eqref{eq:vertices} and \eqref{eq:labels} constraints.

Furthermore, this transformation also preserves the \eqref{eq:cyclic-k} constraint by ensuring that $f'$ achieves the same CAB value as $f$, as proven below:

\begin{center}
\resizebox{!}{!}{
\begin{tabular}{cl}
 & $min(|f'(i) -f'(i')|, n  - |f'(i) - f'(i')|)$ \\
$=$ & $min(|(n + 1 - f(i)) - (n + 1 - f(i'))|, n - |(n + 1 - f(i)) - (n + 1 - f(i'))|)$ \\
$=$ & $min(|f(i') - f(i)|, n - |f(i') - f(i)|)$

\end{tabular}
}
\end{center}

Therefore, in addition to the ILP formulation, we incorporate a symmetry breaking constraint to eliminate the symmetric solution corresponding to the reversed labeling function. By pruning this redundant part of the search space, the proposed encoding effectively reduces the search effort and thereby improves the overall computational performance.

\section{SAT Encoding for Cyclic Ladder constraint} \label{sec:sat_encoding_for_cyclic_ladder_constraint}
This section presents the key contribution of this paper, a compact SAT encoding for CABP that does not encode individual AMO constraints of the \eqref{eq:cyclic-k} constraint separately but instead groups overlapping AMO constraints into blocks and encodes each block independently, while introducing connecting clauses between adjacent blocks to preserve correctness. We first introduce the Cyclic Ladder constraint induced by CABP, then describe its decomposition into blocks and the use of \textit{Sequential Counter} encoding \cite{sinz2005towards} to encode each block with auxiliary variables, followed by the construction of connecting clauses that link adjacent blocks and reconstruct the original set of AMO constraints. Finally, we compare the proposed encoding with direct AMO encodings such as \textit{Pairwise}, \textit{Seq (Sequential Counter)}, \textit{BDD (Binary Decision Diagram)}, and \textit{Card (Cardinality Network}), demonstrating its improved compactness.

\subsection{Definition of Cyclic Ladder constraint}
\label{sec:def_cyclic_ladder}

\begin{definition}
    \label{def:cyclic_ladder}
    Given a sequence of $n$ boolean variables $\Omega = \{x_{1}, x_{2},..., x_{n}\}$ and a number $w$ such that $1 < w \leq n$, a Cyclic Ladder constraint of width $w$ is formulated as follows:
    \[CyclicLadder(\Omega, w)=\bigwedge_{i=1}^{n}\big(\sum_{j=0}^{w-1} x_{t} \leq 1\big), \hspace{1cm} t = \big((i+j-1) \text{ mod } n\big) + 1\]
\end{definition}

A Cyclic Ladder constraint, as defined in Definition~\ref{def:cyclic_ladder}, is a constraint that limits the number of variables assigned the value \textit{true} within consecutive groups of variables arranged along a circular ordering.
For example, in Figure~\ref{fig:cyclic_ladder_constraint_visualization}, nine binary variables $\{x_{1}, \ldots, x_{9}\}$ are arranged in a circle in clockwise order.
Suppose that, for any three consecutive variables on the circle, at most one variable can be assigned \textit{true}, this arrangement generates a sequence of At-most-One (AMO) constraints, as illustrated in Figure~\ref{fig:cyclic_ladder_constraint_9_vars_3_width}.
This sequence of constraints forms a Cyclic Ladder constraint with $n = 9$ and $w = 3$.

\begin{figure}[ht!]
\centering

\begin{minipage}[t!]{0.48\textwidth}
\centering
\resizebox{\textwidth}{!}{
\begin{tikzpicture}[scale=2.5]
  \draw[thick] (0,0) circle (1);
  \def\n{9}
  \def\r{0.6}
  \foreach \i in {1,...,\n} {
    \node[circle,draw,fill=white,minimum size=2*\r cm]
      (C\i) at ({90 - 360/\n * (\i-1)}:1) {$x_{\i}$};
  }
\end{tikzpicture}
}
\caption{\small Graph visualization of the Cyclic Ladder constraint with $n = 9$, $w = 3$.}
\label{fig:cyclic_ladder_constraint_visualization}
\end{minipage}
\hfill
\begin{minipage}[t!]{0.48\textwidth}
\centering
\resizebox{\textwidth}{!}{
\begin{tabular}{llllll}
$x_{1}$ & + & $x_{2}$ & + & $x_{3}$ & $\leq 1 \wedge$ \\[2pt]
$x_{2}$ & + & $x_{3}$ & + & $x_{4}$ & $\leq 1 \wedge$ \\[2pt]
$x_{3}$ & + & $x_{4}$ & + & $x_{5}$ & $\leq 1 \wedge$ \\[2pt]
$x_{4}$ & + & $x_{5}$ & + & $x_{6}$ & $\leq 1 \wedge$ \\[2pt]
$x_{5}$ & + & $x_{6}$ & + & $x_{7}$ & $\leq 1 \wedge$ \\[2pt]
$x_{6}$ & + & $x_{7}$ & + & $x_{8}$ & $\leq 1 \wedge$ \\[2pt]
$x_{7}$ & + & $x_{8}$ & + & $x_{9}$ & $\leq 1 \wedge$ \\[2pt]
$x_{8}$ & + & $x_{9}$ & + & $x_{1}$ & $\leq 1 \wedge$ \\[2pt]
$x_{9}$ & + & $x_{1}$ & + & $x_{2}$ & $\leq 1$
\end{tabular}
}
\vspace{17pt}
\caption{\small ILP formulation of the Cyclic Ladder constraint with $n = 9$, $w = 3$.}
\label{fig:cyclic_ladder_constraint_9_vars_3_width}
\end{minipage}

\end{figure}

\subsection{Decomposition of Cyclic Ladder constraint}

\begin{figure}[ht!]
    \centering
    \includegraphics[width=0.85\linewidth]{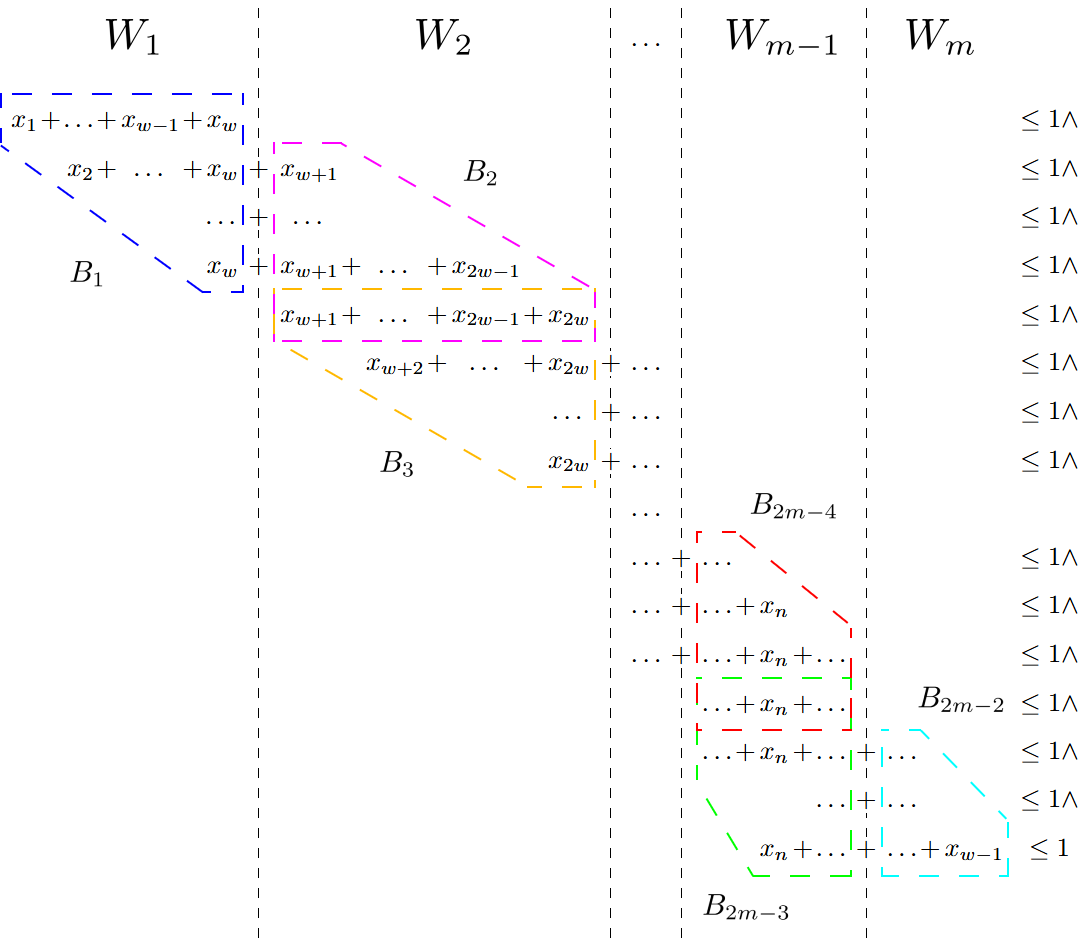}
    \caption{Decomposition of Cyclic Ladder constraint of $n$ variables and width $w$.}
    \label{fig:cyclic_ladder_constraint_decomposition_n_w}

    \vspace{1cm}

    \includegraphics[width=0.85\linewidth]{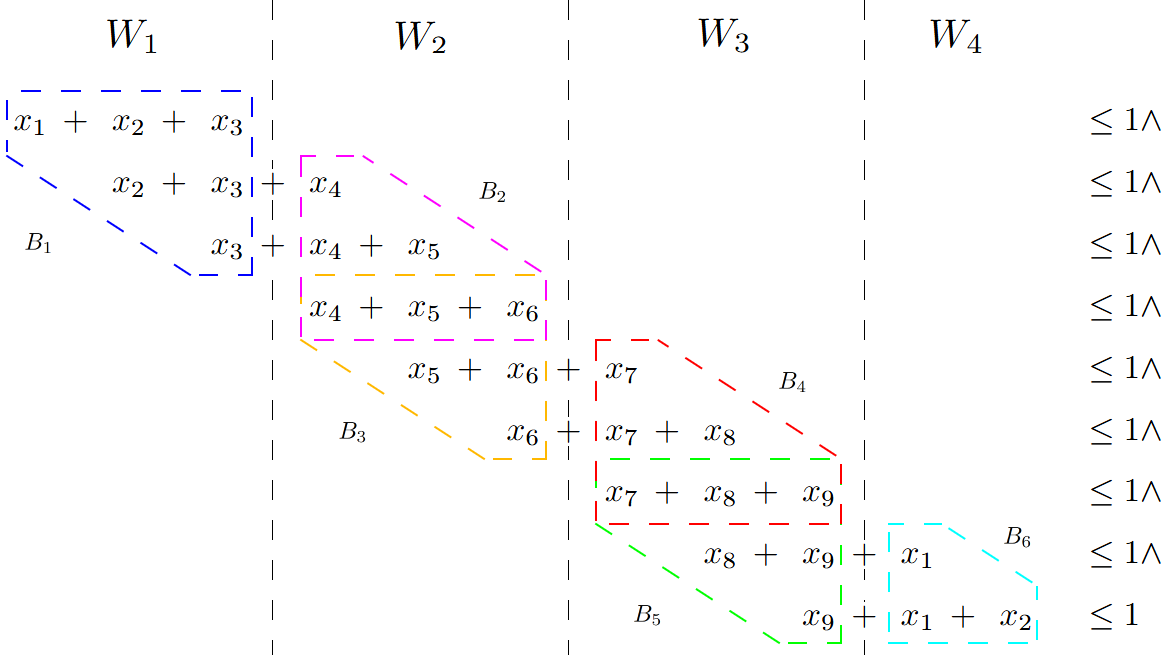}
    \caption{\small Decomposition of Cyclic Ladder constraint of 9 variables and width 3.}
    \label{fig:cyclic_ladder_constraint_decomposition_9_3}
\end{figure}

A Cyclic Ladder constraint over $n$ variables with a width of $w$ can be decomposed into $m = \lceil \frac{n + w - 1}{w} \rceil$ windows $\{W_1, \ldots, W_m\}$, each containing exactly $w$ unique variables, except possibly the last one.
This decomposition simultaneously splits the AMO constraints within the Cyclic Ladder constraint into sub-expressions, which can be reconnected later to reformulate the original constraints by using Proposition~\ref{prop:decompose_AMO_constraint}. These sub-expressions can also be grouped into $(2m-2)$ blocks $\{B_{1}, \ldots, B_{2m-2}\}$ such that all expressions within the same block are mutually nested, as presented in Figure \ref{fig:cyclic_ladder_constraint_decomposition_n_w}.

Except for the last block $B_{2m-2}$, each block consists of exactly $w$ consecutive variables. As a result, each such block naturally induces an AMO constraint within the Cyclic Ladder constraint. Therefore, rather than encoding all AMO constraints directly, we decompose the Cyclic Ladder constraint into blocks and encode only the AMO constraints associated with these blocks, which leads to a substantial reduction in the overall encoding size.

Take the Cyclic Ladder constraint shown in Figure~\ref{fig:cyclic_ladder_constraint_9_vars_3_width} as an example. Its decomposition, illustrated in Figure~\ref{fig:cyclic_ladder_constraint_decomposition_9_3}, partitions the constraint into four windows $\{W_1, \ldots, W_4\}$ and six blocks $\{B_1, \ldots, B_6\}$. Encoding blocks $B_1$ and $B_2$ requires the explicit encoding of two AMO constraints, namely $x_1 + x_2 + x_3 \leq 1$ and $x_4 + x_5 + x_6 \leq 1$. Moreover, these encodings implicitly generate four sub-expressions, including $x_2 + x_3$, $x_3$, $x_4$, and $x_4 + x_5$. By combining these sub-expressions, the encoding of two AMO constraints $x_2 + x_3 + x_4 \leq 1$ and $x_3 + x_4 + x_5 \leq 1$ can be obtained without additional explicit encoding, thereby returning a more compact encoding than encoding each AMO constraint independently.

\subsection{Encoding Decomposed Blocks}
\label{sec:encode-blocks}

Let $x_{i,j}$ be the $j^{th}$ variable of block $i$, according to its variable ordering. 
For example, in Figure~\ref{fig:cyclic_ladder_constraint_decomposition_n_w}, block $B_2$ starts with $x_{w+1}$ and ends with $x_{w+1} + \ldots + x_{2w}$. Therefore, its variable ordering is $x_{w+1} \prec x_{w+2} \prec \ldots \prec x_{2w-1} \prec x_{2w}$. 
As a result, $x_{2,1} \equiv x_{w+1}$, $x_{2,2} \equiv x_{w+2}$, \ldots, $x_{2,w-1} \equiv x_{2w-1}$, and $x_{2,w} \equiv x_{2w}$. 
Similarly, for block $B_3$, whose variable ordering is $x_{2w} \prec x_{2w-1} \prec \ldots \prec x_{w+2} \prec x_{w+1}$, $x_{3,1} \equiv x_{2w}$, $x_{3,2} \equiv x_{2w-1}$, \ldots, $x_{3,w-1} \equiv x_{w+2}$, and $x_{3,w} \equiv x_{w+1}$.

\begin{figure}[ht!]
    \centering
    \includegraphics[width=\linewidth]{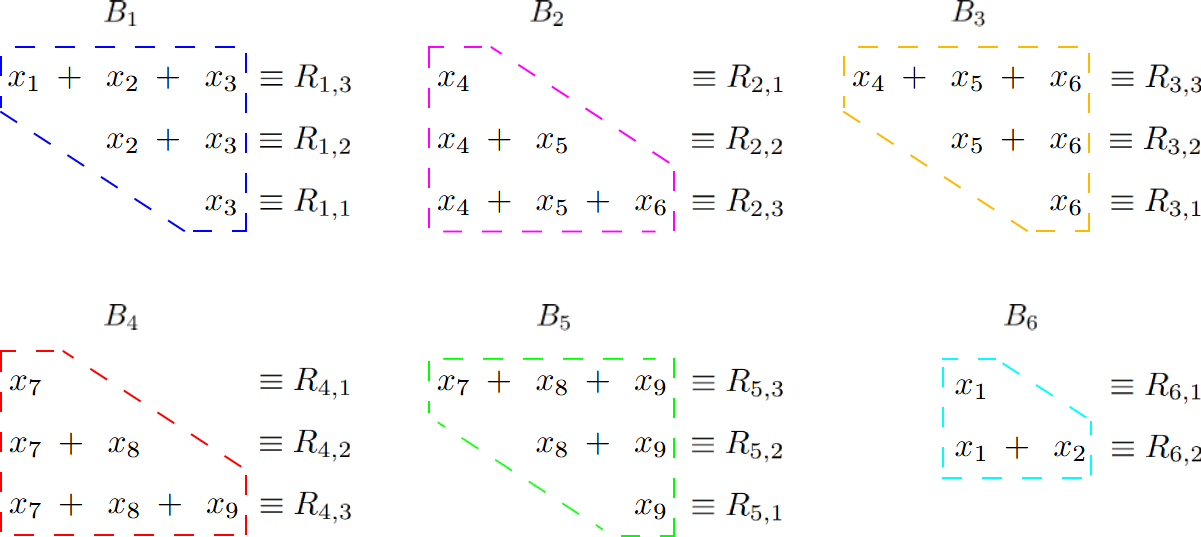}
    \caption{\small Register bit construction of Cyclic Ladder constraint of 9 variables and width 3.}
    \label{fig:cyclic_ladder_constraint_register_bit_9_3}
\end{figure}

Additionally, we construct $R_{i,j}$ as the register bit that indicates the sum of the first $j$ variables in the variable ordering of block $i$. $R_{i,j}$ is set to $true$ if and only if there is exactly one among the first $j$ variables of block $i$ assigned $true$, i.e., $\sum_{j'=1}^{j} x_{i,j'} = 1$. Conversely, $R_{i,j}$ is $false$ if and only if none of the first $j$ variables of block $i$ is assigned $true$, i.e., $\sum_{j'=1}^{j} x_{i,j'} \leq 0$. Figure \ref{fig:cyclic_ladder_constraint_register_bit_9_3} illustrates the register bit construction for the six blocks obtained from the decomposition of the Cyclic Ladder constraint in Figure \ref{fig:cyclic_ladder_constraint_decomposition_9_3}.

With the definitions of $x_{i,j}$ and $R_{i,j}$, we apply the formulation proposed by Hieu et al. \cite{Truong2025} to encode the decomposed blocks. 
This formulation is based on the Sequential Counter encoding \cite{sinz2005towards} and consists of four formulas \eqref{eq:block_encoding_1}, \eqref{eq:block_encoding_2}, \eqref{eq:block_encoding_3}, and \eqref{eq:block_encoding_4}. 
Formula \eqref{eq:block_encoding_1} sets $R_{i,j}$ to $true$ when the $j^{th}$ variable $x_{i,j}$ is $true$. 
Formula \eqref{eq:block_encoding_2} sets $R_{i,j}$ to $true$ when the cumulative sum of the first $j-1$ variables, represented by $R_{i,j-1}$, is $true$. 
Formula~\eqref{eq:block_encoding_3} sets $R_{i,j}$ to $false$ when all of the first $j$ variables are $false$ (i.e., both $x_{i,j}$ and $R_{i,j-1}$ are $false$). 
Finally, Formula~\eqref{eq:block_encoding_4} ensures that if the $j^{th}$ variable $x_{i,j}$ is $true$, then the cumulative sum of the first $j-1$ variables, represented by $R_{i,j-1}$, must be $false$, thereby guaranteeing that at most one variable can be assigned $true$.

\begin{minipage}[t]{0.4\textwidth}
\begin{equation}
    \label{eq:block_encoding_1}
    \bigwedge_{j=2}^{w}x_{i,j}\to R_{i,j}
\end{equation}

\begin{equation}
    \label{eq:block_encoding_2}
    \bigwedge_{j=2}^{w} R_{i,j-1} \to R_{i,j}
\end{equation}
\end{minipage}
\hfill
\begin{minipage}[t]{0.56\textwidth}
\begin{equation}
    \label{eq:block_encoding_3}
    \bigwedge_{j=2}^{w}\neg x_{i,j}\wedge\neg R_{i,j-1}\to \neg R_{i,j}
\end{equation}

\begin{equation}
    \label{eq:block_encoding_4}
    \bigwedge_{j=2}^{w} x_{i,j} \to \neg R_{i,j-1}
\end{equation}
\end{minipage}

Applying these formulas to encode block $B_1$ in the Cyclic Ladder constraint in Figure \ref{fig:cyclic_ladder_constraint_decomposition_9_3}, the following clauses are obtained:

\resizebox{0.95\textwidth}{!}{
\begin{tabular}{cccrclcrcl}
\multirow{2}{*}{\eqref{eq:block_encoding_1}} & $\bigwedge_{j=2}^{3}$ & \multirow{2}{*}{$\iff$} & $x_{1,2}$ & $\to$ & $R_{1,2}$ & \multirow{2}{*}{$\iff$} & $x_{2}$ & $\to$ & $R_{1,2}$ \\
 & $x_{1,j}\to R_{1,j}$ &  & $x_{1,3}$ & $\to$ & $R_{1,3}$ &  & $x_{1}$ & $\to$ & $R_{1,3}$ \\
 &  &  &  &  &  &  &  &  &  \\
\multirow{2}{*}{\eqref{eq:block_encoding_2}} & $\bigwedge_{j=2}^{3}$ & \multirow{2}{*}{$\iff$} & $R_{1,1}$ & $\to$ & $R_{1,2}$ & \multirow{2}{*}{$\iff$} & $x_{3}$ & $\to$ & $R_{1,2}$ \\
 & $R_{1,j-1} \to R_{1,j}$ &  & $R_{1,2}$ & $\to$ & $R_{1,3}$ &  & $R_{1,2}$ & $\to$ & $R_{1,3}$ \\
 &  &  &  &  &  &  &  &  &  \\
\multirow{2}{*}{\eqref{eq:block_encoding_3}} & $\bigwedge_{j=2}^{3}$ & \multirow{2}{*}{$\iff$} & $\neg x_{1,2} \wedge \neg R_{1,1}$ & $\to$ & $\neg R_{1,2}$ & \multirow{2}{*}{$\iff$} & $\neg x_{2} \wedge \neg x_{3}$ & $\to$ & $\neg R_{1,2}$ \\
 & $\neg x_{1,j}\wedge\neg R_{1,j-1}\to \neg R_{1,j}$ &  & $\neg x_{1,3} \wedge \neg R_{1,2}$ & $\to$ & $\neg R_{1,3}$ &  & $\neg x_{1} \wedge \neg R_{1,2}$ & $\to$ & $\neg R_{1,3}$ \\
 &  &  &  &  &  &  &  &  &  \\
\multirow{2}{*}{\eqref{eq:block_encoding_4}} & $\bigwedge_{j=2}^{3}$ & \multirow{2}{*}{$\iff$} & $x_{1,2}$ & $\to$ & $\neg R_{1,1}$ & \multirow{2}{*}{$\iff$} & $x_{2}$ & $\to$ & $\neg x_{3}$ \\
 & $x_{1,j} \to \neg R_{1,j-1}$ &  & $x_{1,3}$ & $\to$ & $\neg R_{1,2}$ &  & $x_{1}$ & $\to$ & $\neg R_{1,2}$
\end{tabular}
}

Similarly, encoding block $B_2$ in the Cyclic Ladder constraint in Figure \ref{fig:cyclic_ladder_constraint_decomposition_9_3} results in the following clauses:

\resizebox{0.95\textwidth}{!}{
\begin{tabular}{cccrclcrcl}
\multirow{2}{*}{\eqref{eq:block_encoding_1}} & $\bigwedge_{j=2}^{3}$ & \multirow{2}{*}{$\iff$} & $x_{2,2}$ & $\to$ & $R_{2,2}$ & \multirow{2}{*}{$\iff$} & $x_{5}$ & $\to$ & $R_{2,2}$ \\
 & $x_{2,j}\to R_{2,j}$ &  & $x_{2,3}$ & $\to$ & $R_{2,3}$ &  & $x_{6}$ & $\to$ & $R_{2,3}$ \\
 &  &  &  &  &  &  &  &  &  \\
\multirow{2}{*}{\eqref{eq:block_encoding_2}} & $\bigwedge_{j=2}^{3}$ & \multirow{2}{*}{$\iff$} & $R_{2,1}$ & $\to$ & $R_{2,2}$ & \multirow{2}{*}{$\iff$} & $x_{4}$ & $\to$ & $R_{2,2}$ \\
 & $R_{2,j-1} \to R_{2,j}$ &  & $R_{2,2}$ & $\to$ & $R_{2,3}$ &  & $R_{2,2}$ & $\to$ & $R_{2,3}$ \\
 &  &  &  &  &  &  &  &  &  \\
\multirow{2}{*}{\eqref{eq:block_encoding_3}} & $\bigwedge_{j=2}^{3}$ & \multirow{2}{*}{$\iff$} & $\neg x_{2,2} \wedge \neg R_{2,1}$ & $\to$ & $\neg R_{2,2}$ & \multirow{2}{*}{$\iff$} & $\neg x_{5} \wedge \neg x_{4}$ & $\to$ & $\neg R_{2,2}$ \\
 & $\neg x_{2,j}\wedge\neg R_{2,j-1}\to \neg R_{2,j}$ &  & $\neg x_{2,3} \wedge \neg R_{2,2}$ & $\to$ & $\neg R_{2,3}$ &  & $\neg x_{6} \wedge \neg R_{2,2}$ & $\to$ & $\neg R_{2,3}$ \\
 &  &  &  &  &  &  &  &  &  \\
\multirow{2}{*}{\eqref{eq:block_encoding_4}} & $\bigwedge_{j=2}^{3}$ & \multirow{2}{*}{$\iff$} & $x_{2,2}$ & $\to$ & $\neg R_{2,1}$ & \multirow{2}{*}{$\iff$} & $x_{5}$ & $\to$ & $\neg x_{4}$ \\
 & $x_{2,j} \to \neg R_{2,j-1}$ &  & $x_{2,3}$ & $\to$ & $\neg R_{2,2}$ &  & $x_{6}$ & $\to$ & $\neg R_{2,2}$
\end{tabular}
}

When encoding blocks that share the same set of variables, i.e., blocks $B_2$ and $B_3$, the constraint enforced by formula \eqref{eq:block_encoding_4}, which ensures that at most one variable is assigned the value $true$, only needs to be applied once. Consequently, since the encoding of block $B_2$ already includes formula \eqref{eq:block_encoding_4}, block $B_3$ is encoded using the three formulas \eqref{eq:block_encoding_1}, \eqref{eq:block_encoding_2}, and \eqref{eq:block_encoding_3}, as follows:

\resizebox{0.95\textwidth}{!}{
\begin{tabular}{cccrclcrcl}
\multirow{2}{*}{\eqref{eq:block_encoding_1}} & $\bigwedge_{j=2}^{3}$ & \multirow{2}{*}{$\iff$} & $x_{3,2}$ & $\to$ & $R_{3,2}$ & \multirow{2}{*}{$\iff$} & $x_{5}$ & $\to$ & $R_{3,2}$ \\
 & $x_{3,j}\to R_{3,j}$ &  & $x_{3,3}$ & $\to$ & $R_{3,3}$ &  & $x_{4}$ & $\to$ & $R_{3,3}$ \\
 &  &  &  &  &  &  &  &  &  \\
\multirow{2}{*}{\eqref{eq:block_encoding_2}} & $\bigwedge_{j=2}^{3}$ & \multirow{2}{*}{$\iff$} & $R_{3,1}$ & $\to$ & $R_{3,2}$ & \multirow{2}{*}{$\iff$} & $x_{6}$ & $\to$ & $R_{3,2}$ \\
 & $R_{3,j-1} \to R_{3,j}$ &  & $R_{3,2}$ & $\to$ & $R_{3,3}$ &  & $R_{3,2}$ & $\to$ & $R_{3,3}$ \\
 &  &  &  &  &  &  &  &  &  \\
\multirow{2}{*}{\eqref{eq:block_encoding_3}} & $\bigwedge_{j=2}^{3}$ & \multirow{2}{*}{$\iff$} & $\neg x_{3,2} \wedge \neg R_{3,1}$ & $\to$ & $\neg R_{3,2}$ & \multirow{2}{*}{$\iff$} & $\neg x_{5} \wedge \neg x_{6}$ & $\to$ & $\neg R_{3,2}$ \\
 & $\neg x_{3,j}\wedge\neg R_{3,j-1}\to \neg R_{3,j}$ &  & $\neg x_{3,3} \wedge \neg R_{3,2}$ & $\to$ & $\neg R_{3,3}$ &  & $\neg x_{4} \wedge \neg R_{3,2}$ & $\to$ & $\neg R_{3,3}$ \\
 &  &  &  &  &  &  &  &  &  \\
\end{tabular}
}

In cases where the last block contains fewer than $w$ variables, the final block is encoded by applying the four formulas \eqref{eq:block_encoding_1} to \eqref{eq:block_encoding_4} with the actual number of variables in that block instead of $w$. For example, the last block $B_6$ contains only two variables, i.e., $x_1$ and $x_2$. By replacing the value of $w$ with 2, the encoding of this block is given as follows:

\resizebox{0.95\textwidth}{!}{
\begin{tabular}{cccrclcrcl}
\multirow{2}{*}{\eqref{eq:block_encoding_1}} & $\bigwedge_{j=2}^{2}$ & \multirow{2}{*}{$\iff$} & \multirow{2}{*}{$x_{6,2}$} & \multirow{2}{*}{$\to$} & \multirow{2}{*}{$R_{6,2}$} & \multirow{2}{*}{$\iff$} & \multirow{2}{*}{$x_{2}$} & \multirow{2}{*}{$\to$} & \multirow{2}{*}{$R_{6,2}$} \\
 & $x_{6,j}\to R_{6,j}$ &  &  &  &  &  &  &  &  \\
 &  &  &  &  &  &  &  &  &  \\
\multirow{2}{*}{\eqref{eq:block_encoding_2}} & $\bigwedge_{j=2}^{2}$ & \multirow{2}{*}{$\iff$} & \multirow{2}{*}{$R_{6,1}$} & \multirow{2}{*}{$\to$} & \multirow{2}{*}{$R_{6,2}$} & \multirow{2}{*}{$\iff$} & \multirow{2}{*}{$x_{1}$} & \multirow{2}{*}{$\to$} & \multirow{2}{*}{$R_{6,2}$} \\
 & $R_{6,j-1} \to R_{6,j}$ &  &  &  &  &  &  &  &  \\
 &  &  &  &  &  &  &  &  &  \\
\multirow{2}{*}{\eqref{eq:block_encoding_3}} & $\bigwedge_{j=2}^{2}$ & \multirow{2}{*}{$\iff$} & \multirow{2}{*}{$\neg x_{6,2} \wedge \neg R_{6,1}$} & \multirow{2}{*}{$\to$} & \multirow{2}{*}{$\neg R_{6,2}$} & \multirow{2}{*}{$\iff$} & \multirow{2}{*}{$\neg x_{2} \wedge \neg x_{1}$} & \multirow{2}{*}{$\to$} & \multirow{2}{*}{$\neg R_{6,2}$} \\
 & $\neg x_{6,j}\wedge\neg R_{6,j-1}\to \neg R_{6,j}$ &  &  &  &  &  &  &  &  \\
 &  &  &  &  &  &  &  &  &  \\
\multirow{2}{*}{\eqref{eq:block_encoding_4}} & $\bigwedge_{j=2}^{2}$ & \multirow{2}{*}{$\iff$} & \multirow{2}{*}{$x_{6,2}$} & \multirow{2}{*}{$\to$} & \multirow{2}{*}{$\neg R_{6,1}$} & \multirow{2}{*}{$\iff$} & \multirow{2}{*}{$x_{2}$} & \multirow{2}{*}{$\to$} & \multirow{2}{*}{$\neg x_{1}$} \\
 & $x_{6,j} \to \neg R_{2,j-1}$ &  &  &  &  &  &  &  & 
\end{tabular}
}

\subsection{Connecting Decomposed Blocks}

After encoding the decomposed blocks, we construct connecting clauses between adjacent blocks to guaranty the correctness of the original Cyclic Ladder constraint.
This process employs Proposition \ref{prop:decompose_AMO_constraint}, in which the corresponding expressions of each block are linked to form the AMO constraints within the Cyclic Ladder constraint.

For example, connecting the two blocks $B_1$ and $B_2$ in the Cyclic Ladder constraint shown in Figure~\ref{fig:cyclic_ladder_constraint_decomposition_9_3} requires linking $x_2 + x_3$ with $x_4$ to obtain $x_2 + x_3 + x_4$, and linking $x_3$ with $x_4 + x_5$ to obtain $x_3 + x_4 + x_5$. By applying Proposition~\ref{prop:decompose_AMO_constraint}, these links are constructed as follows:

\resizebox{0.95\textwidth}{!}{

    \begin{tabular}{ccccc}
    \multirow{2}{*}{$(x_{2} + x_{3} + x_{4} \leq 1)$} & \multirow{2}{*}{$\overset{\text{Prop. \ref{prop:decompose_AMO_constraint}}}{\equiv}$} & $(x_{2} + x_{3} \leq 1) \wedge (x_{4} \leq 1)$ & \multirow{2}{*}{$\equiv$} & $(x_{2} + x_{3} \leq 1) \wedge (x_{4} \leq 1)$ \\
     &  & $\wedge (x_{2} + x_{3} \leq 0 \vee x_{4} \leq 0)$ \rule{0pt}{12pt} &  & $\wedge( \neg R_{1,2} \vee \neg x_4)$ \\
    \end{tabular}
}

\resizebox{0.95\textwidth}{!}{

    \begin{tabular}{ccccc}
    \multirow{2}{*}{$(x_{3} + x_{4} + x_{5} \leq 1)$} & \multirow{2}{*}{$\overset{\text{Prop. \ref{prop:decompose_AMO_constraint}}}{\equiv}$} & $(x_{3} \leq 1) \wedge (x_{4} + x_{5} \leq 1)$ & \multirow{2}{*}{$\equiv$} & $(x_{3} \leq 1) \wedge (x_{4} + x_{5} \leq 1)$ \\
     &  & $\wedge (x_{3} \leq 0 \vee x_{4} + x_{5} \leq 0)$ \rule{0pt}{12pt} &  & $\wedge(\neg x_{3} \vee \neg R_{2,2})$
    \end{tabular}
}

Similarly, connecting the remaining block, i.e., $\{B_3, B_4\}$ and $\{B_5, B_6\}$, requires the application of the following clauses:

\resizebox{0.95\textwidth}{!}{
\begin{tabular}{ccccc}
\multirow{2}{*}{$(x_{5} + x_{6} + x_{7} \leq 1)$} & \multirow{2}{*}{$\overset{\text{Prop. \ref{prop:decompose_AMO_constraint}}}{\equiv}$} & $(x_{5} + x_{6} \leq 1) \wedge (x_{7} \leq 1)$ & \multirow{2}{*}{$\equiv$} & $(x_{5} + x_{6} \leq 1) \wedge (x_{7} \leq 1)$ \\
 &  & $\wedge (x_{5} + x_{6} \leq 0 \vee x_{7} \leq 0)$ &  & $\wedge (\neg R_{3,2} \vee \neg x_{7})$ \rule{0pt}{12pt} \\
\end{tabular}
}

\resizebox{0.95\textwidth}{!}{
\begin{tabular}{ccccc}
\multirow{2}{*}{$(x_{6} + x_{7} + x_{8} \leq 1)$} & \multirow{2}{*}{$\overset{\text{Prop. \ref{prop:decompose_AMO_constraint}}}{\equiv}$} & $(x_{6} \leq 1) \wedge (x_{7} + x_{8} \leq 1)$ & \multirow{2}{*}{$\equiv$} & $(x_{6} \leq 1) \wedge (x_{7} + x_{8} \leq 1)$ \\
 &  & $\wedge (x_{6} \leq 0 \vee x_{7} + x_{8} \leq 0)$ &  & $\wedge(\neg x_{6} \vee \neg R_{4,2})$ \rule{0pt}{12pt} \\
\end{tabular}
}

\resizebox{0.95\textwidth}{!}{
\begin{tabular}{ccccc}
\multirow{2}{*}{$(x_{8} + x_{9} + x_{1} \leq 1)$} & \multirow{2}{*}{$\overset{\text{Prop. \ref{prop:decompose_AMO_constraint}}}{\equiv}$} & $(x_{8} + x_{9} \leq 1) \wedge (x_{1} \leq 1)$ & \multirow{2}{*}{$\equiv$} & $(x_{8} + x_{9} \leq 1) \wedge (x_{1} \leq 1)$ \\
 &  & $\wedge (x_{8} + x_{9} \leq 0 \vee x_{1} \leq 0)$ \rule{0pt}{12pt} &  & $\wedge (\neg R_{5,2} \vee \neg x_{1})$\\
\end{tabular}
}

\resizebox{0.95\textwidth}{!}{
\begin{tabular}{ccccc}
\multirow{2}{*}{$(x_{9} + x_{1} + x_{2} \leq 1)$} & \multirow{2}{*}{$\overset{\text{Prop. \ref{prop:decompose_AMO_constraint}}}{\equiv}$} & $(x_{9} \leq 1) \wedge (x_{1} + x_{2} \leq 1)$ & \multirow{2}{*}{$\equiv$} & $(x_{9} \leq 1) \wedge (x_{1} + x_{2} \leq 1)$ \\
 &  & $\wedge (x_{9} \leq 0 \vee x_{1} + x_{2} \leq 0)$ \rule{0pt}{12pt} &  & $\wedge (\neg x_9 \vee \neg R_{6,2})$
\end{tabular}
}

\subsection{Comparison with Direct SAT Encodings}

This section compares the proposed encoding for the Cyclic Ladder constraint with several SAT encodings that encode each AMO constraint directly, including \textit{Pairwise}, \textit{Sequential Counter} (\textit{Seq}) \cite{sinz2005towards}, \textit{Binary Decision Diagram} (\textit{BDD}) \cite{abio2012new}, \textit{Cardinality Network} (\textit{Card}) \cite{asin2009cardinality, asin2011cardinality}, and \textit{Product} \cite{chen2010new}. Table~\ref{tab:compare-cyclic-ladder-constraint} summarizes this comparison by reporting, for each encoding, the number of auxiliary variables, the number of clauses, and the asymptotic clause-space complexity required to encode a Cyclic Ladder constraint with $n$ variables and width $w$. Moreover, since the structure of the Cyclic Ladder constraint, as well as its block decomposition and block encoding formulation, are closely related to a type of constraint studied by Hieu et al.~\cite{Truong2025}, this comparison can be referred to the one presented in~\cite{Truong2025} for further details.

\begin{table*}[ht!]
\caption{\small Size of SAT encoding for a Cyclic Ladder constraint over $n$ variables with width $w$.}
\label{tab:compare-cyclic-ladder-constraint}
\centering
\resizebox{\textwidth}{!}{
\begin{tabular}{|c|c|c|c|c|}
\hline
\multirow{2}{*}{\textbf{Encoding}} & \multirow{2}{*}{\textbf{Technique}} & \multirow{2}{*}{\textbf{Auxiliary variables}} & \multirow{2}{*}{\textbf{Clauses}} & \multirow{2}{*}{\textbf{Complexity}} \\
 &  &  &  &  \\ \hline
\multirow{2}{*}{\textit{Pairwise}} & \multirow{10}{*}{\begin{tabular}[c]{@{}c@{}}Encode \\ AMO constraints \\ directly\end{tabular}} & \multirow{2}{*}{0} & \multirow{2}{*}{$\frac{(w-1)w}{2} + (n-1)(w-1)$} & \multirow{2}{*}{$\mathcal{O}(n^2)$} \\
 &  &  &  &  \\ \cline{1-1} \cline{3-5} 
\multirow{2}{*}{\textit{Seq}} &  & \multirow{2}{*}{$n(w-2)$} & \multirow{2}{*}{$n(4w-7)$} & \multirow{2}{*}{$\mathcal{O}(n^2)$} \\
 &  &  &  &  \\ \cline{1-1} \cline{3-5} 
\multirow{2}{*}{\textit{BDD}} &  & \multirow{2}{*}{$2n(w-1)$} & \multirow{2}{*}{$4n(w-1)$} & \multirow{2}{*}{$\mathcal{O}(n^2)$} \\
 &  &  &  &  \\ \cline{1-1} \cline{3-5} 
\multirow{2}{*}{\textit{Card}} &  & \multirow{2}{*}{$\mathcal{O}(n^2)$} & \multirow{2}{*}{$\mathcal{O}(n^2)$} & \multirow{2}{*}{$\mathcal{O}(n^2)$} \\
 &  &  &  &  \\ \cline{1-1} \cline{3-5} 
\multirow{2}{*}{\textit{Product}} &  & \multirow{2}{*}{$n(2\sqrt{w} + \mathcal{O}\sqrt[4]{w})$} & \multirow{2}{*}{$n(2w+4\sqrt{w}+\mathcal{O}\sqrt[4]{w})$} & \multirow{2}{*}{$\mathcal{O}(n^2)$} \\
 &  &  &  &  \\ \hline
Proposed & \multirow{2}{*}{\begin{tabular}[c]{@{}c@{}}Encode set of \\ AMO constraints\end{tabular}} & \multirow{2}{*}{$2mw -3m -2w +4$} & \multirow{2}{*}{$8mw - 8m - 7w + 7$} & \multirow{2}{*}{$\mathcal{O}(n)$} \\
encoding &  &  &  &  \\ \hline
\end{tabular}
}
\end{table*}

Assume that all $(2m-2)$ decomposed blocks have the same size, although the last block is typically smaller. In the proposed encoding, $m$ blocks are encoded using all four formulas \eqref{eq:block_encoding_1}–\eqref{eq:block_encoding_4}, whereas the remaining $(m-2)$ blocks are encoded using only the three formulas \eqref{eq:block_encoding_1}–\eqref{eq:block_encoding_3} (as explained by the encoding of block $B_3$ provided in Section~\ref{sec:encode-blocks}). In addition, connecting the $(2m-2)$ blocks requires $(m-1)$ connections, each contributing $(w-1)$ clauses. As each formula \eqref{eq:block_encoding_1}–\eqref{eq:block_encoding_4} introduces $(w-1)$ clauses, the total number of clauses used to encode a Cyclic Ladder constraint with $n$ variables and width $w$ is $4(w-1)m + 3(w-1)(m-2) + (m-1)(w-1) = 8mw -8m -7w + 7$. Since $m = \lceil \frac{n + w - 1}{w} \rceil$, the proposed encoding achieves the clause-space complexity of $\mathcal{O}(n)$.

In terms of auxiliary variables, the proposed encoding introduces $(w-1)$ such variables per block. However, when a block shares the same variable set with a previously encoded block (e.g., block $B_2$ and $B_3$ in Figure~\ref{fig:cyclic_ladder_constraint_decomposition_9_3}), one auxiliary variable can be reused instead of introducing a new one (i.e., $R_{3,3}$ reuses $R_{2,3}$). This reduces the number of auxiliary variables for such blocks from $(w-1)$ to $(w-2)$. Consequently, the total number of auxiliary variables required by the proposed encoding is $m(w-1) + (m-2)(w-2) = 2mw - 3m -2w + 4$.

Table~\ref{tab:compare-cyclic-ladder-constraint} highlights the compactness and scalability of the proposed encoding in comparison to five direct encodings for the Cyclic Ladder constraint. In particular, the proposed encoding requires substantially fewer auxiliary variables and clauses than \textit{Seq}, \textit{BDD}, \textit{Card}, and \textit{Product}. Additionally, these encodings incur quadratic growth in the size of the SAT formulation, either in the number of auxiliary variables, the number of clauses, or both. By contrast, the proposed encoding exploits the structure of the Cyclic Ladder constraint and encodes a set of AMO constraints simultaneously, resulting in a linear-size encoding with respect to $n$. Besides, although the \textit{Pairwise} encoding avoids the introduction of auxiliary variables, this direct encoding still generates a quadratic number of clauses, which limits its scalability on large instances.

\section{Experimental Setup}\label{sec4}

This section presents the experimental setup used to evaluate the performance of the SAT-based approach, referred to as SAT-CAB (SAT for Cyclic Antibandwidth). First, the implementation details of SAT-CAB are outlined. Subsequently, the datasets utilized in the experiments are introduced, followed by a description of the experimental environment and the evaluation metrics used to assess performance.

\subsection{Implementation of SAT-CAB}

\subsubsection{Implementation of Parallel SAT Solving}

To better exploit available computational resources and reduce overall solving time, SAT-CAB implements a parallel SAT-based framework, which is an extended version of the iterative SAT-based procedure presented in Section~\ref{sec:iterative-sat-solving}. The objective of this parallel framework is similar to that of the iterative approach, which is to identify a CAB value \(k'\) for which a feasible labeling exists, while no feasible labeling exists for \(k' + 1\). 

The key difference between the iterative and parallel approaches lies in the adopted solving strategy. Rather than examining candidate values one by one, the parallel framework employs a multi-processing strategy in which multiple ILP models, each associated with a specific value of \(k\), are solved simultaneously. In addition, to further accelerate the search, the candidate values can be prioritized according to a predefined ordering instead of being explored in a strictly sequential order. In SAT-CAB, this prioritization is guided by a binary pruning strategy, which organizes candidate values using tree traversal algorithms such as breadth-first search and depth-first search.

The parallel framework implemented in SAT-CAB works as described in Algorithm~\ref{alg:sat_cab_parallel}. The procedure begins by initializing the search queue and launching the search for feasible labelings of several candidate CAB values with the highest priority. When SAT-CAB determines that a candidate CAB value is feasible (resp. infeasible), it updates the corresponding results and infers the feasibility of other related candidate CAB values. Subsequently, SAT-CAB continues to process the remaining candidate values to determine their feasibility, repeating the above steps until the feasibility of all candidate values has either been explicitly proven or inferred.

\begin{algorithm}[ht!]
\caption{Parallel SAT solving for CABP}
\label{alg:sat_cab_parallel}
\begin{algorithmic}[1]
\Require Graph $G=(V,E)$, SAT solver $\mathcal{S}$, processes $\mathcal{P}$, bounds $LB,UB$
\Ensure Optimal CAB value $k_{opt}$ and labeling

\State Initialize priority search queue $\mathcal{Q}$, $k_{SAT}\gets LB-1$, $k_{UNSAT}\gets UB+1$

\For{$i=1$ \textbf{to} $\min(|\mathcal{P}|,UB-LB+1)$}
    \State Launch process to solve ILP model with $k\!\leftarrow\!\textsc{Pop}(\mathcal{Q}\cap[k_{SAT},k_{UNSAT}])$
\EndFor

\While{active processes exist}
    \State Wait for any process PID$(k')$ to finish
    \If{PID$(k')$ is SAT}
        \State $k_{SAT} \gets k'$
        \State terminate PID$(k'')$ with $k''<k'$
    \Else
        \State $k_{UNSAT} \gets k'$
        \State terminate PID$(k'')$ with $k''>k'$
    \EndIf
    \State Store labeling if SAT
    \While{free process exists \textbf{and} $\mathcal{Q}\cap[k_{SAT},k_{UNSAT}]\neq\emptyset$}
        \State Launch process to solve ILP model with $k\!\leftarrow\!\textsc{Pop}(\mathcal{Q}\cap[k_{SAT},k_{UNSAT}])$
    \EndWhile
\EndWhile

\State \textbf{return} $k_{SAT}$ as $k_{opt}$ and corresponding labeling
\end{algorithmic}
\end{algorithm}

\subsubsection{Implementation of ILP Model}

The encoding of the ILP model in SAT-CAB proceeds as follows. For each edge, the ~\eqref{eq:cyclic-k} constraint is first decomposed into two Cyclic Ladder constraints. Each Cyclic Ladder constraint is then encoded using the encoding proposed in Section ~\ref{sec:sat_encoding_for_cyclic_ladder_constraint}. The auxiliary variables introduced during the encoding are subsequently employed to guarantee the correctness of decomposing the original AMO constraint into two AMO constraints (as stated in Proposition \ref{prop:decompose_AMO_constraint}).

Next, since the encoding of the Cyclic Ladder constraint already introduces auxiliary variables to represent sub-expressions, the constraint~\eqref{eq:labels} can reuse these variables to obtain an efficient encoding. 
For example, in the Cyclic Ladder constraint illustrated in Figure~\ref{fig:cyclic_ladder_constraint_decomposition_9_3}, three auxiliary variables are introduced to represent the sub-expressions \(\{x_1 + x_2 + x_3\}\), \(\{x_4 + x_5 + x_6\}\), and \(\{x_7 + x_8 + x_9\}\) (see Figure~\ref{fig:cyclic_ladder_constraint_register_bit_9_3}). 
Consequently, the corresponding constraint~\eqref{eq:labels}, namely \(x_1 + \cdots + x_9 = 1\), can be encoded using these three auxiliary variables instead of the nine original variables \(\{x_1,\ldots,x_9\}\), thereby significantly reducing the encoding size.

Inspired by the implementation of the encoding for the Antibandwidth problem given by Fazekas et al.~\cite{fazekas2020duplex}, the constraint~\eqref{eq:vertices} is implemented by combining the \textit{Product} encoding~\cite{chen2010new} with the \textit{Sequential Counter} encoding~\cite{sinz2005towards}. 
The \textit{Product} encoding is first employed to restructure the original variables into a two-dimensional array. 
Specifically, \(n\) variables are mapped to a \(p \times q\) matrix with \(p \cdot q = n\), thereby replacing the original \(n\) variables with only \(p + q\) auxiliary variables. 
Subsequently, the \textit{Sequential Counter} encoding is applied along each row and each column of the resulting two-dimensional array to enforce the exactly one constraint.

It is worth noting that, as described in Section~\ref{sec:symmetry-breaking}, for any labeling function \(f\) that satisfies the ILP model with \(k = k'\), there exists a corresponding labeling function \(f'\) that also satisfies the same ILP model. Consequently, we apply a symmetry breaking technique along with the ILP formulation in order to reduce the search space. This technique is inspired by the work of Sinnl~\cite{sinnl2021note}, which enforces that the vertex with the maximum degree in the graph is assigned a label no greater than $\frac{|V|}{2}$. In the case of multiple vertices attaining the maximum degree, the vertex with the smallest index is selected.

\subsection{Dataset}

The experiments consider seven sets of problems with a total of 175 instances, including 20 instances of \textit{three-dimensional mesh} (\textit{3D mesh}), 20 instances of \textit{double stars}, 7 instances of \textit{hypercube}, 40 instances of \textit{caterpillar}, 24 instances of \textit{complete binary tree} (\textit{CBT}), 24 instances of \textit{Harwell-Boeing graph}, and 40 instances of \textit{random connected graph}. 
These datasets are taken from the work of Lozano et al. \cite{lozano2013hybrid} and currently available at \url{https://grafo.etsii.urjc.es/optsicom/cabp.html}.

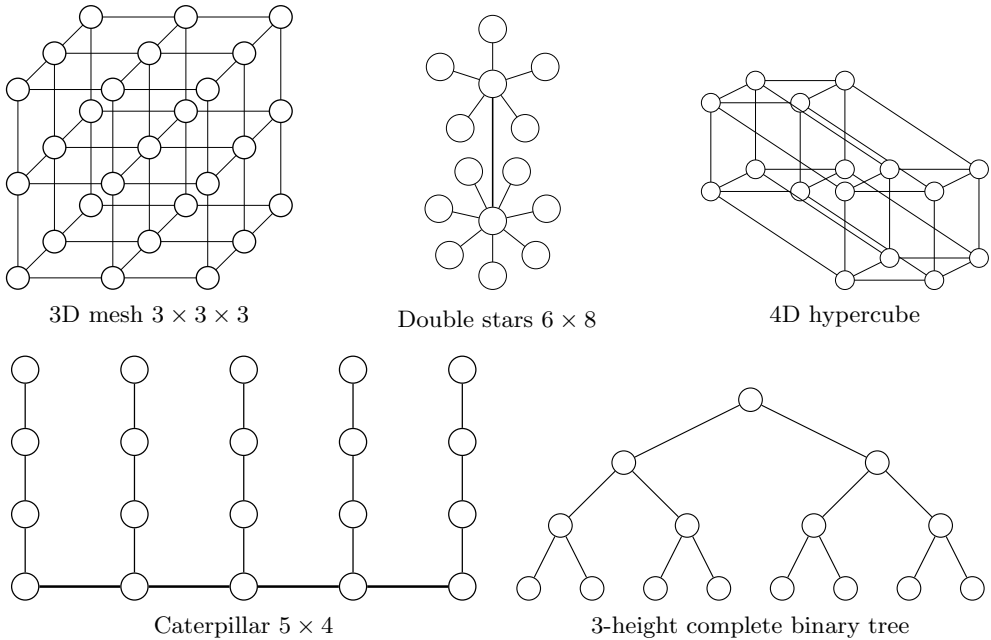
\begin{figure}[ht!]
\centering

\begin{minipage}[t]{0.3\textwidth}
\centering
\resizebox{\textwidth}{!}{
\begin{tikzpicture}[scale=2]
  \foreach \i in {0,...,2}{
    \foreach \j in {0,...,2}{
      \foreach \k in {0,...,2}{
        \coordinate (P\i\j\k) at (\i,\j,\k);
      }
    }
  }
  \foreach \i in {0,1}{
    \foreach \j in {0,1,2}{
      \foreach \k in {0,1,2}{
        \pgfmathtruncatemacro{\ip}{\i+1}
        \draw (P\i\j\k) -- (P\ip\j\k);
      }
    }
  }
  \foreach \i in {0,1,2}{
    \foreach \j in {0,1}{
      \foreach \k in {0,1,2}{
        \pgfmathtruncatemacro{\jp}{\j+1}
        \draw (P\i\j\k) -- (P\i\jp\k);
      }
    }
  }
  \foreach \i in {0,1,2}{
    \foreach \j in {0,1,2}{
      \foreach \k in {0,1}{
        \pgfmathtruncatemacro{\kp}{\k+1}
        \draw (P\i\j\k) -- (P\i\j\kp);
      }
    }
  }
  \foreach \i in {0,...,2}{
    \foreach \j in {0,...,2}{
      \foreach \k in {0,...,2}{
        \filldraw[fill=white,draw=black,thick] (P\i\j\k) circle (0.12);
      }
    }
  }
\end{tikzpicture}
}
\vspace{2pt}
{\small 3D mesh $3\times3\times3$}
\end{minipage}
\hfill
\begin{minipage}[t]{0.3\textwidth}
\centering
\resizebox{0.5\textwidth}{!}{
\begin{tikzpicture}[every node/.style={circle,draw=black,fill=white,minimum size=5mm}]
    \def\nA{5}
    \def\nB{7}
    \def\dist{2.5} 
    \def\radius{1} 
    \node[] (C1) at (0,0) {};
    \node[] (C2) at (0,-\dist) {};
    \foreach \i in {1,...,\nA} {
        \coordinate (tmpA\i) at ($(C1) + ({90-360/\nA*(\i-1)}:\radius)$);
        \draw[thin] (C1) -- (tmpA\i);        
        \node (A\i) at (tmpA\i) {};        
    }
    \foreach \i in {1,...,\nB} {
        \coordinate (tmpB\i) at ($(C2) + ({-90+360/\nB*(\i-1)}:\radius)$);
        \draw[thin] (C2) -- (tmpB\i);
        \node (B\i) at (tmpB\i) {};
    }
    \draw[thick] (C1) -- (C2);
\end{tikzpicture}
}
\vspace{2pt} \\
{\small Double stars $6\times8$}
\end{minipage}
\hfill
\begin{minipage}[t]{0.3\textwidth}
\centering
\resizebox{\textwidth}{!}{
\begin{tikzpicture}[scale=2, line join=round]
  \newcommand{\cube}[1]{
    \coordinate (#1A) at (0,0);
    \coordinate (#1B) at (1,0);
    \coordinate (#1C) at (1,1);
    \coordinate (#1D) at (0,1);
    \coordinate (#1E) at (0.5,0.25);
    \coordinate (#1F) at (1.5,0.25);
    \coordinate (#1G) at (1.5,1.25);
    \coordinate (#1H) at (0.5,1.25);
    \draw (#1A)--(#1B)--(#1C)--(#1D)--cycle;
    \draw (#1E)--(#1F)--(#1G)--(#1H)--cycle;
    \draw (#1A)--(#1E) (#1B)--(#1F) (#1C)--(#1G) (#1D)--(#1H);
  }
  \cube{C}
  \begin{scope}[shift={(1.5,-1)}]
    \cube{D}
  \end{scope}
  \foreach \p in {A,B,C,D,E,F,G,H}{
    \draw (C\p)--(D\p);
    \filldraw[fill=white] (C\p) circle(3pt);
    \filldraw[fill=white] (D\p) circle(3pt);
  }
\end{tikzpicture}
}
\vspace{2pt}
{\small 4D hypercube}
\end{minipage}

\vspace{8pt}

\begin{minipage}[t]{0.49\textwidth}
\centering
\resizebox{\textwidth}{!}{
\begin{tikzpicture}[
  every node/.style={circle,draw,fill=white,inner sep=3pt},
  spine/.style={thick}
]
  \foreach \i in {1,...,5}{
    \node (s\i) at ({1.2*(\i-1)},0) {};
    \ifnum\i>1 \draw[spine] (s\the\numexpr\i-1\relax)--(s\i); \fi
    \foreach \j in {1,2,3}{
      \node (l\i\j) at ({1.2*(\i-1)}, {0.8*\j}) {};
      \ifnum\j=1 \draw (s\i)--(l\i\j); \else \draw (l\i\the\numexpr\j-1\relax)--(l\i\j); \fi
    }
  }
\end{tikzpicture}
}
\vspace{2pt}
{\small Caterpillar $5\times4$}
\end{minipage}
\hfill
\begin{minipage}[t]{0.49\textwidth}
\centering
\resizebox{\textwidth}{!}{
    \begin{tikzpicture}[
      level distance=0.8cm,
      level 1/.style={sibling distance=3.2cm},
      level 2/.style={sibling distance=1.6cm},
      level 3/.style={sibling distance=0.8cm},
      every node/.style={circle,draw,fill=white,inner sep=3pt}
    ]
    \node {}
      child {node {}
        child {node {}
          child {node {}}
          child {node {}}
        }
        child {node {}
          child {node {}}
          child {node {}}
        }
      }
      child {node {}
        child {node {}
          child {node {}}
          child {node {}}
        }
        child {node {}
          child {node {}}
          child {node {}}
        }
      };
    \end{tikzpicture}
}
\vspace{2pt}
{\small 3-height complete binary tree}
\end{minipage}

\caption{Shapes of several structured graphs.}
\label{fig:shape_of_graphs}
\end{figure}

Among these datasets, five correspond to structured graph families: \textit{3D mesh}, \textit{double stars}, \textit{hypercube}, \textit{caterpillar}, and \textit{complete binary tree}. 
The structures of these graph types are illustrated in Figure~\ref{fig:shape_of_graphs}.
For the remaining datasets, the \textit{Harwell–Boeing graph} originates from the Harwell–Boeing Sparse Matrix Collection\footnote{\url{https://math.nist.gov/MatrixMarket/collections/hb.html}}, a set of standard test matrices derived from problems in linear systems, least squares, and eigenvalue computations, while the \textit{random connected graph} is obtained by connecting vertices randomly according to a specified probability.
A detailed description of these datasets and their corresponding CAB bounds is provided in the Appendix~\ref{secA1}.

\subsection{Experimental Environment and Metrics} \label{sec-experiment-setup}

The source code of SAT-CAB is implemented in C++.
To ensure reproducibility, all experiments are executed on a virtual machine hosted on Google Cloud Platform\footnote{\url{https://console.cloud.google.com/}}, running Debian 12 on a 64-bit architecture. 
This machine is configured with the machine type e2-highmem-8, equipped with 4 cores, 8 virtual CPUs, 64 gigabytes (GB) of memory, and no GPUs. All experiments employ the CaDiCaL SAT solver \cite{BiereSATRace2019}, version~2.1.3, with a memory limit of 30~GB per instance.
Moreover, SAT-CAB is executed under two parallel configurations, using 4 and 8 processes, respectively, and the best result among the two runs is reported.

The experiments employ four primary performance metrics under a given time limit: (1) the number of solved instances (\textit{\#Solved}), defined as those for which a labeling is found for at least one CAB value within the given bounds; (2) the number of best CAB values achieved among all approaches (\textit{\#Best}); (3) the number of optimal CAB values identified (\textit{\#Optimal}); and (4) the number of new CAB values discovered (\textit{\#New}), determined by comparing the obtained results with those reported by previously studied approaches.
Note that the time consumption of each approach is measured over the entire algorithm, including all stages of instance modeling (e.g., formulation encoding) and the solution search (e.g., invoking solver to search for a solution).

\section{Experimental Results} \label{sec4-2-2}

This section evaluates the performance of the proposed SAT-CAB method by comparing it against three previously studied metaheuristic methods, namely MACAB \cite{bansal2011memetic}, HABC-CAB \cite{lozano2013hybrid}, and MS-GVNS \cite{cavero2022general}, across six benchmark datasets, including \textit{3D mesh}, \textit{hypercube}, \textit{double stars}, \textit{caterpillar}, \textit{complete binary tree} (\textit{CBT}), and \textit{Harwell-Boeing graph}. 
Since the source code of MACAB, HABC-CAB, and MS-GVNS is not publicly available, the comparison relies on the report of Cavero et al. \cite{cavero2022general}, accessible at \url{https://grafo.etsii.urjc.es/optsicom/cab/results.xlsx}. 
Note that, in this report, HABC-CAB and MS-GVNS were run by Cavero et al. \cite{cavero2022general} on their local machine, while the results of MACAB were taken from the experiments of Lozano et al. \cite{lozano2013hybrid}.
Additionally, the comparison on \textit{Harwell-Boeing graph} also includes the results of the HACO-CAB obtained from the work of Sundar \cite{sundar2019hybrid}. 
The hardware configurations used in the experiments for MACAB, HABC-CAB, HACO-CAB, and MS-GVNS are summarized in Table~\ref{tab:macab-habcab-hacocab-configurations}.

\begin{table*}[ht!]
    \caption{\small Harware configurations of MACAB, HABC-CAB, HACO-CAB and MS-GVNS.}
    \label{tab:macab-habcab-hacocab-configurations}
    \centering
    \resizebox{!}{!}{
    \begin{tabular}{l|ccc}
    \hline
     & Processor & RAM & Operator System \\ \hline
    MACAB & 3.2 GHz Intel\textsuperscript{\textregistered} Core\textsuperscript{\textregistered} i7 & 12GB & Fedora\textsuperscript{\textregistered} Linux V15 OS \\
    HABC-CAB & 2.2 GHz Intel\textsuperscript{\textregistered} Core\textsuperscript{\textregistered} i7 & \multicolumn{1}{l}{16GB} & - \\
    HACO-CAB & 3.1 GHz Intel\textsuperscript{\textregistered} Core\textsuperscript{\textregistered} i5 & 4GB & Linux-based \\
    MS-GVNS & 2.2 GHz Intel\textsuperscript{\textregistered} Core\textsuperscript{\textregistered} i7 & \multicolumn{1}{l}{16GB} & - \\ \hline
    \end{tabular}
    }
\end{table*}

In addition to these metaheuristic approaches, SAT-CAB is also compared with Constraint Programming (CP) and Mixed Integer Programming (MIP) approaches, including CPLEX\textsubscript{CP}, CPLEX\textsubscript{MIP}, and Gurobi. 
Particularly, CPLEX\textsubscript{CP} and CPLEX\textsubscript{MIP} respectively utilize the CP and MIP API provided by IBM ILOG CPLEX Optimization Studio\footnote{\url{https://www.ibm.com/products/ilog-cplex-optimization-studio}, Version 22.1.1.}, while Gurobi employs the MIP API provided by Gurobi Optimizer\footnote{\url{https://www.gurobi.com/}, Version 11.0.3.}. 
It is worth noting that both IBM ILOG CPLEX Optimization Studio and Gurobi Optimizer are widely used commercial solvers in practical applications.

Since the \textit{random connected graph} instances used in previous studies \cite{bansal2011memetic, lozano2013hybrid, cavero2022general} could not be accessed for download, the evaluation of this dataset was instead conducted on 40 instances obtained from \url{https://grafo.etsii.urjc.es/optsicom/cabp.html}, which are also provided by Lozano et al.~\cite{lozano2013hybrid}.
Furthermore, because of the unavailability of the source code for MACAB, HABC-CAB, HACO-CAB, and MS-GVNS, it was not possible to rerun these metaheuristic approaches in a unified experimental setting. 
As a consequence, the results of these approaches are excluded from the comparison on the \textit{random connected graph} dataset. 
Therefore, the analysis on this dataset focuses exclusively on SAT-CAB, CPLEX\textsubscript{CP}, CPLEX\textsubscript{MIP}, and Gurobi, allowing for a fair comparison among approaches that can be executed under identical hardware and software configurations.

\subsection{Evaluation on general graphs}

\subsubsection{The Harwell-Boeing Graph}

Table~\ref{tab:result_harwell_boeing_to_150s} reports the experimental results on the \textit{Harwell-Boeing graph} dataset, presenting the CAB values obtained by each approach across all 24 instances under a 150-second time limit.
The best CAB value achieved for each instance across all approaches is highlighted in bold.
Any CAB value that is proven to be optimal is additionally marked with an asterisk (*). 
If an approach fails to find a feasible labeling for all CAB values of an instance within the given bounds under the time limits (resp. memory limits), the corresponding entry in the table is marked as TO (resp. MO).

Along with the main results, Table~\ref{tab:result_harwell_boeing_to_150s} also includes four additional rows, namely \textit{\#Solved}, \textit{\#Best}, \textit{\#Optimal}, and \textit{\#New}. The first three rows \textit{\#Solved}, \textit{\#Best}, \textit{\#Optimal} indicate, respectively, the number of instances solved, the number of instances for which the approach attains the best result among all the approaches, and the number of instances for which the optimal CAB value is found.
The last row reports the number of instances for which a new CAB value is identified by CPLEX\textsubscript{CP}, CPLEX\textsubscript{MIP}, Gurobi, and SAT-CAB, compared with MACAB, HABC-CAB, HACO-CAB, and MS-GVNS.

\begin{table*}[ht!]
    \caption{\small Experiment results of Harwell-Boeing graph instances.}
    \label{tab:result_harwell_boeing_to_150s}
    \centering
    \resizebox{\textwidth}{!}{
    \begin{tabular}{l|cccc|cccc}
    \hline
    \multirow{2}{*}{Instance} & \multicolumn{4}{c|}{\textbf{TO = 150s}\footnotemark} & \multicolumn{4}{c}{\textbf{TO = 150s}} \\ \cline{2-9} 
     & \textbf{MACAB} & \textbf{HABC-CAB} & \textbf{HACO-CAB} & \textbf{MS-GVNS} & \textbf{CPLEX\textsubscript{CP}} & \textbf{CPLEX\textsubscript{MIP}} & \textbf{Gurobi} & \textbf{SAT-CAB} \\ \hline
    A-pores\_1 & 6 & 6 & 6 & 6 & \textbf{6*} & \textbf{6*} & \textbf{6*} & \textbf{6*} \\
    B-ibm32 & 6 & 8 & 8 & 8 & \textbf{8*} & \textbf{8*} & \textbf{8*} & \textbf{8*} \\
    C-bcspwr01 & 9 & 13 & 13 & 13 & \textbf{13*} & \textbf{13*} & \textbf{13*} & \textbf{13*} \\
    D-bcsstk01 & 6 & 7 & 8 & 8 & \textbf{8*} & \textbf{8*} & \textbf{8*} & \textbf{8*} \\
    E-bcspwr02 & 12 & 16 & 16 & 16 & \textbf{16*} & \textbf{16*} & \textbf{16*} & \textbf{16*} \\
    F-curtis54 & 7 & 10 & 10 & 10 & \textbf{10*} & \textbf{10*} & \textbf{10*} & \textbf{10*} \\
    G-will57 & 7 & 11 & 11 & 11 & \textbf{11*} & \textbf{11*} & \textbf{11*} & \textbf{11*} \\
    H-impcol\_b & 3 & 7 & 7 & 7 & 7 & \textbf{7*} & \textbf{7*} & \textbf{7*} \\
    I-ash85 & 13 & 19 & 18 & 20 & \textbf{21*} & 19 & 20 & \textbf{21*} \\
    J-nos4 & 22 & 30 & 30 & 31 & \textbf{32*} & 30 & 25 & \textbf{32*} \\
    K-dwt\_\_234 & 22 & 43 & 44 & 45 & \textbf{46} & 44 & 36 & \textbf{46} \\
    L-bcspwr03 & 18 & 29 & 29 & 29 & \textbf{29*} & 29 & 29 & \textbf{29*} \\ \hline \hline
    M-bcsstk07 & 12 & 29 & 27 & 30 & 27 & TO & 16 & \textbf{32} \\
    N-bcsstk06 & 12 & 29 & 27 & 30 & 27 & TO & 16 & \textbf{32} \\
    O-impcol\_d & 30 & 84 & 81 & 84 & 87 & TO & TO & \textbf{93} \\
    P-can\_\_445 & 36 & 73 & 72 & 73 & 65 & TO & 41 & \textbf{79} \\
    Q-494\_bus & 53 & 163 & 162 & 164 & \textbf{164*} & TO & TO & \textbf{164*} \\
    R-dwt\_\_503 & 14 & 48 & 42 & 45 & 46 & TO & TO & \textbf{53} \\
    S-sherman4 & \textbf{258} & 256 & 254 & 257 & 193 & TO & TO & 136 \\
    T-dwt\_\_592 & 33 & \textbf{97} & 96 & 95 & 80 & TO & TO & 76 \\
    U-662\_bus & 58 & 165 & 165 & 165 & 165 & TO & TO & \textbf{165*} \\
    V-nos6 & \textbf{328} & 325 & 325 & 325 & 231 & TO & TO & TO \\
    W-685\_bus & 16 & 114 & 114 & 114 & 114 & TO & TO & \textbf{114*} \\
    X-can\_\_715 & 21 & 98 & 96 & \textbf{99} & 90 & TO & TO & 72 \\ \hline
    \textbf{\#Solved} & \textbf{24} & \textbf{24} & \textbf{24} & \textbf{24} & \textbf{24} & 12 & 15 & 23 \\
    \textbf{\#Best} & 2 & 1 & 0 & 1 & 12 & 8 & 8 & \textbf{20} \\
    \textbf{\#Optimal} & 0 & 0 & 0 & 0 & 11 & 8 & 8 & \textbf{14} \\
    \textbf{\#New} & - & - & - & - & 4 & 0 & 0 & \textbf{8} \\ \hline
    \end{tabular}
    }
\end{table*}

Based on the results presented in Table~\ref{tab:result_harwell_boeing_to_150s}, the SAT-CAB method demonstrates superior performance compared to the benchmark approaches. Among the 12 small-sized instances (from \textit{A-pores\_1} to \textit{L-bcspwr03}), SAT-CAB found the optimal CAB values for 11 instances, surpassing CPLEX\textsubscript{CP}, CPLEX\textsubscript{MIP}, and Gurobi. Moreover, when compared to metaheuristic search methods, SAT-CAB also achieves better results on several instances. In particular, SAT-CAB discovers new CAB values for three instances \textit{I-ash85}, \textit{J-nos4}, and \textit{K-dwt\_\_234}, with the two instances \textit{I-ash85} and \textit{J-nos4} also proved to be optimal.

\footnotetext{Results taken from the work of Sundar \cite{sundar2019hybrid} and Cavero et al. \cite{cavero2022general}.}

For the 12 large-sized \textit{Harwell-Boeing graph} instances, SAT-CAB obtained the optimal CAB values for three instances, namely \textit{Q-494\_bus}, \textit{U-662\_bus}, and \textit{W-685\_bus}, while among the other approaches, only CPLEX\textsubscript{CP} was able to find the optimal value for instance \textit{Q-494\_bus}. In addition, SAT-CAB discovered new CAB values for five other instances, including \textit{M-bcsstk07}, \textit{N-bcsstk06}, \textit{O-impcol\_d}, \textit{P-can\_\_445}, and \textit{R-dwt\_\_503}.

Overall, considering all 24 instances of the Harwell–Boeing dataset under a 150-second time limit, SAT-CAB clearly outperforms other approaches. Furthermore, while the other approaches are able to prove optimal CAB values for at most 11 instances and achieve the best CAB values for no more than 12 instances, SAT-CAB significantly exceeds these results. Specifically, SAT-CAB proves optimality for 14 instances and attains the best CAB values on 20 instances, thereby demonstrating its clear superiority in terms of solution quality and robustness.

To further extend the CAB results for the \textit{Harwell-Boeing graph} dataset, additional experiments were conducted on instances for which the optimal CAB values had not yet been found, increasing the time limit from 150 seconds to 1800 seconds. 
The experimental results obtained by the four methods, including CPLEX\textsubscript{CP}, CPLEX\textsubscript{MIP}, Gurobi, and SAT-CAB, are summarized in Table~\ref{tab:result_harwell_boeing_to_1800s}.
Under the extended 1800-second time limit, SAT-CAB continues to demonstrate superior performance over the CP and MIP-based approaches by discovering improved CAB values for eight out of the ten instances that had not yet reached the optimal value. Moreover, among these eight instances, SAT-CAB is able to prove the optimality of the new CAB values obtained for the two instances \textit{O-impcol\_d} and \textit{R-dwt\_\_503}.

\begin{table*}[ht!]
    \caption{Experiment results of non-optimal {Harwell-Boeing graph} instances with time limit of 1800 seconds.}
    \label{tab:result_harwell_boeing_to_1800s}
    \centering
    \resizebox{\textwidth}{!}{
    \begin{tabular}{l|ccccc|cccc}
    \hline
    \multirow{2}{*}{Instance} & \multicolumn{5}{c|}{\textbf{TO = 150s}} & \multicolumn{4}{c}{\textbf{TO = 1800s}} \\ \cline{2-10} 
     & \textbf{MACAB} & \textbf{HABC-CAB} & \textbf{HACO-CAB} & \textbf{MS-GVNS} & \textbf{SAT-CAB} & \textbf{CPLEX\textsubscript{CP}} & \textbf{CPLEX\textsubscript{MIP}} & \textbf{Gurobi} & \textbf{SAT-CAB} \\ \hline
    K-dwt\_\_234 & 22 & 43 & 44 & 45 & \textbf{46} & \textbf{46} & 45 & 43 & \textbf{46} \\
    M-bcsstk07 & 12 & 29 & 27 & 30 & 32 & 29 & 21 & 22 & \textbf{33} \\
    N-bcsstk06 & 12 & 29 & 27 & 30 & 32 & 29 & 21 & 22 & \textbf{33} \\
    O-impcol\_d & 30 & 84 & 81 & 84 & 93 & 104 & 53 & 61 & \textbf{105*} \\
    P-can\_\_445 & 36 & 73 & 72 & 73 & 79 & 76 & 43 & 49 & \textbf{87} \\
    R-dwt\_\_503 & 14 & 48 & 42 & 45 & 53 & 50 & 27 & 35 & \textbf{62*} \\
    S-sherman4 & \textbf{258} & 256 & 254 & 257 & 136 & 246 & TO & TO & 227 \\
    T-dwt\_\_592 & 33 & 97 & 96 & 95 & 76 & 87 & TO & 55 & \textbf{113} \\
    V-nos6 & \textbf{328} & 325 & 325 & 325 & TO & 298 & TO & TO & 272 \\
    X-can\_\_715 & 21 & 98 & 96 & 99 & 72 & 97 & TO & 60 & \textbf{101} \\ \hline
    \textbf{\#Solved} & \textbf{10} & \textbf{10} & \textbf{10} & \textbf{10} & 9 & \textbf{10} & 6 & 8 & \textbf{10} \\
    \textbf{\#Best} & 2 & 0 & 0 & 0 & 1 & 1 & 0 & 0 & \textbf{8} \\
    \textbf{\#Optimal} & 0 & 0 & 0 & 0 & 0 & 0 & 0 & 0 & \textbf{2} \\
    \textbf{\#New} & - & - & - & - & 6 & 3 & 0 & 0 & \textbf{8} \\ \hline
    \end{tabular}
    }
\end{table*}

\subsubsection{The random connected graph}

Table~\ref{tab:result_random_connected} presents the experimental results of four approaches, including CPLEX\textsubscript{CP}, CPLEX\textsubscript{MIP}, Gurobi, and SAT-CAB, on 40 \textit{random connected graph} instances\footnote{These instances are available at \url{https://grafo.etsii.urjc.es/optsicom/cabp.html}}. In this table, the columns $|V|$ and $|E|$ denote the numbers of vertices and edges of each instance. 
The CAB values that are proven to be optimal for each instance are reported in the \textit{\#Optimal} column. For instances whose optimal values have not yet been determined in our experiments,  the reported lower bound ($\#LB$) is the largest value proven SAT by SAT-CAB, while the upper bound ($\#UB$) corresponds to the smallest value whose successor ($\#UB+1$) is proven UNSAT.

\begin{table*}[ht!]
    \caption{Experimental results of random connected graph instances.}
    \label{tab:result_random_connected}
    \centering
    \resizebox{\textwidth}{!}{
        \begin{tabular}{lcc|ccccccc}
        \hline
        \multirow{2}{*}{\textbf{Inst.}} & \multirow{2}{*}{\textbf{$|$V$|$}} & \multirow{2}{*}{\textbf{$|$E$|$}} & \multicolumn{7}{c}{\textbf{TO = 1800s}} \\ \cline{4-10} 
         &  &  & \textbf{CPLEX\textsubscript{CP}} & \textbf{CPLEX\textsubscript{MIP}} & \textbf{Gurobi} & \multicolumn{1}{c|}{\textbf{SAT-CAB}} & \textbf{\#Optimal} & \textbf{\#LB} & \textbf{\#UB} \\ \hline
        p1 & 100 & 200 & \textbf{32*} & 31 & 31 & \multicolumn{1}{c|}{\textbf{32*}} & 32 & - & - \\
        p2 & 100 & 200 & \textbf{31} & 30 & \textbf{31} & \multicolumn{1}{c|}{\textbf{31}} & - & 31 & 32 \\
        p3 & 100 & 200 & \textbf{31} & 30 & \textbf{31} & \multicolumn{1}{c|}{\textbf{31}} & - & 31 & 32 \\
        p4 & 100 & 200 & 31 & 31 & 30 & \multicolumn{1}{c|}{\textbf{31*}} & 31 & - & - \\
        p5 & 100 & 200 & 31 & 31 & 30 & \multicolumn{1}{c|}{\textbf{31*}} & 31 & - & - \\
        p6 & 100 & 200 & 31 & 30 & 30 & \multicolumn{1}{c|}{\textbf{31*}} & 31 & - & - \\
        p7 & 100 & 200 & \textbf{31} & 30 & \textbf{31} & \multicolumn{1}{c|}{\textbf{31}} & - & 31 & 32 \\
        p8 & 100 & 200 & 31 & 31 & 31 & \multicolumn{1}{c|}{\textbf{31*}} & 31 & - & - \\
        p9 & 100 & 200 & 31 & 31 & 31 & \multicolumn{1}{c|}{\textbf{31*}} & 31 & - & - \\
        p10 & 100 & 200 & \textbf{32*} & 31 & 31 & \multicolumn{1}{c|}{\textbf{32*}} & 32 & - & - \\
        p11 & 100 & 600 & 16 & 15 & 14 & \multicolumn{1}{c|}{\textbf{17}} & - & 17 & 19 \\
        p12 & 100 & 600 & 16 & 15 & 15 & \multicolumn{1}{c|}{\textbf{17}} & - & 17 & 19 \\
        p13 & 100 & 600 & 16 & 15 & 15 & \multicolumn{1}{c|}{\textbf{17}} & - & 17 & 19 \\
        p14 & 100 & 600 & 15 & 14 & 14 & \multicolumn{1}{c|}{\textbf{17}} & - & 17 & 19 \\
        p15 & 100 & 600 & 16 & 15 & 15 & \multicolumn{1}{c|}{\textbf{17}} & - & 17 & 19 \\
        p16 & 100 & 600 & 16 & 15 & 15 & \multicolumn{1}{c|}{\textbf{17}} & - & 17 & 19 \\
        p17 & 100 & 600 & 15 & 16 & 14 & \multicolumn{1}{c|}{\textbf{17}} & - & 17 & 19 \\
        p18 & 100 & 600 & \textbf{16} & 15 & 15 & \multicolumn{1}{c|}{\textbf{16}} & - & 16 & 21 \\
        p19 & 100 & 600 & 15 & 15 & 14 & \multicolumn{1}{c|}{\textbf{17}} & - & 17 & 20 \\
        p20 & 100 & 600 & 15 & 15 & 14 & \multicolumn{1}{c|}{\textbf{17}} & - & 17 & 20 \\
        p21 & 200 & 400 & 63 & 58 & 41 & \multicolumn{1}{c|}{\textbf{64}} & - & 64 & 65 \\
        p22 & 200 & 400 & 63 & 57 & 39 & \multicolumn{1}{c|}{\textbf{64}} & - & 64 & 65 \\
        p23 & 200 & 400 & 63 & 57 & 39 & \multicolumn{1}{c|}{\textbf{64}} & - & 64 & 65 \\
        p24 & 200 & 400 & \textbf{63} & 57 & 40 & \multicolumn{1}{c|}{\textbf{63}} & - & 63 & 65 \\
        p25 & 200 & 400 & \textbf{63} & 58 & 38 & \multicolumn{1}{c|}{\textbf{63}} & - & 63 & 65 \\
        p26 & 200 & 400 & \textbf{63} & 58 & 38 & \multicolumn{1}{c|}{\textbf{63}} & - & 63 & 65 \\
        p27 & 200 & 400 & 64 & 58 & 41 & \multicolumn{1}{c|}{\textbf{65*}} & 65 & - & - \\
        p28 & 200 & 400 & \textbf{64} & 57 & 41 & \multicolumn{1}{c|}{\textbf{64}} & - & 64 & 65 \\
        p29 & 200 & 400 & \textbf{65} & 59 & 42 & \multicolumn{1}{c|}{\textbf{65}} & - & 65 & 66 \\
        p30 & 200 & 400 & \textbf{65} & 59 & 42 & \multicolumn{1}{c|}{\textbf{65}} & - & 65 & 66 \\
        p31 & 200 & 2000 & 19 & 20 & 17 & \multicolumn{1}{c|}{\textbf{22}} & - & 22 & 43 \\
        p32 & 200 & 2000 & 19 & 18 & 18 & \multicolumn{1}{c|}{\textbf{22}} & - & 22 & 43 \\
        p33 & 200 & 2000 & 19 & 18 & 18 & \multicolumn{1}{c|}{\textbf{22}} & - & 22 & 43 \\
        p34 & 200 & 2000 & 19 & 20 & 17 & \multicolumn{1}{c|}{\textbf{22}} & - & 22 & 43 \\
        p35 & 200 & 2000 & 19 & 19 & 18 & \multicolumn{1}{c|}{\textbf{22}} & - & 22 & 43 \\
        p36 & 200 & 2000 & 19 & 19 & 17 & \multicolumn{1}{c|}{\textbf{22}} & - & 22 & 43 \\
        p37 & 200 & 2000 & 19 & 19 & 17 & \multicolumn{1}{c|}{\textbf{22}} & - & 22 & 43 \\
        p38 & 200 & 2000 & 19 & 14 & 18 & \multicolumn{1}{c|}{\textbf{22}} & - & 22 & 43 \\
        p39 & 200 & 2000 & 19 & 18 & 18 & \multicolumn{1}{c|}{\textbf{22}} & - & 22 & 43 \\
        p40 & 200 & 2000 & 19 & 18 & 18 & \multicolumn{1}{c|}{\textbf{22}} & - & 22 & 43 \\ \hline
        \multicolumn{3}{l|}{\textbf{\#Solved}} & \textbf{40} & \textbf{40} & \textbf{40} & \multicolumn{1}{c|}{\textbf{40}} &  &  &  \\
        \multicolumn{3}{l|}{\textbf{\#Best}} & 12 & 0 & 3 & \multicolumn{1}{c|}{\textbf{40}} &  &  &  \\
        \multicolumn{3}{l|}{\textbf{\#Optimal}} & 2 & 0 & 0 & \multicolumn{1}{c|}{\textbf{8}} &  &  &  \\ \hline
        \end{tabular}
    }
\end{table*}

The results presented in Table~\ref{tab:result_random_connected} demonstrate that SAT-CAB outperforms the three compared approaches. It achieves the best CAB value on all 40 instances, whereas CPLEX\textsubscript{CP} does so on only 12 instances and Gurobi on just 3. Furthermore, SAT-CAB is able to prove optimality for 8 instances. This result is substantially stronger than that of CPLEX\textsubscript{CP}, which identifies optimal solutions for only 2 instances, while neither CPLEX\textsubscript{MIP} nor Gurobi proves optimality for any instance.

In addition, for instances whose optimality has not been determined, the $\#LB$ and $\#UB$ columns indicate that SAT-CAB provides very tight bounds on the CAB value. In most cases, the gap between the lower and upper bounds is only one or two units, especially for medium-sized instances with $|V|=100$ and $|E|=600$, or $|V|=200$ and $|E|=400$. Even for larger instances ($|V|=200$, $|E|=2000$), although the bounds are not as tight as those for the smaller instances, SAT-CAB is still able to narrow the remaining search space. Overall, this result shows that SAT-CAB is effective not only in finding good solutions and proving optimality for \textit{random connected graph} dataset, but also in providing reliable lower and upper bounds that clearly indicate how close the current solutions are to the optimal values, making the comparison with other approaches more informative.

\subsection{Evaluation on structured graph instances}

Table~\ref{tab:summary-structured-graph-results} summarizes the experimental results of MACAB, HABC-CAB, MS-GVNS, CPLEX\textsubscript{CP}, CPLEX\textsubscript{MIP}, Gurobi, and SAT-CAB on five structured graph datasets, namely \textit{3D mesh}, \textit{double stars}, \textit{hypercube}, \textit{caterpillar}, and \textit{complete binary tree}. For each dataset, the table reports the number of instances solved, the number of instances for which each approach achieves the best result among all the approaches, the number of instances whose optimality is proven, and the number of instances for which new CAB values are discovered. The detailed results of each approach at the instance-level can be found in Appendix~\ref{sec:appendB}.

\begin{table*}[ht!]
    \caption{Experimental results of structured graph datasets.}
    \label{tab:summary-structured-graph-results}
    \centering
    \resizebox{\textwidth}{!}{
    \begin{tabular}{c|l|ccccccc}
    \hline
    \textbf{Dataset} & \multicolumn{1}{c|}{\textbf{Approach}} & \textbf{MACAB} & \textbf{HABC-CAB} & \textbf{MS-GVNS} & \textbf{CPLEX\textsubscript{CP}} & \textbf{CPLEX\textsubscript{MIP}} & \textbf{Gurobi} & \textbf{SAT-CAB} \\ \hline
     & \#Solved & \textbf{20} & \textbf{20} & \textbf{20} & \textbf{20} & 7 & 10 & 8 \\
    3D mesh & \#Best & \textbf{15} & 1 & 2 & 4 & 2 & 2 & 5 \\
    (20 instances) & \#Optimal & 0 & 0 & 0 & 2 & 2 & 2 & \textbf{3} \\
     & \#New & - & - & - & 0 & 0 & 0 & \textbf{2} \\ \hline
     & \#Solved & \textbf{20} & \textbf{20} & \textbf{20} & \textbf{20} & \textbf{20} & \textbf{20} & \textbf{20} \\
    Double stars & \#Best & 0 & 0 & 0 & 0 & \textbf{20} & \textbf{20} & 3 \\
    (20 instances) & \#Optimal & 0 & 0 & 0 & 0 & \textbf{20} & \textbf{20} & 3 \\
     & \#New & - & - & - & 0 & 0 & 0 & 0 \\ \hline
     & \#Solved & \textbf{7} & \textbf{7} & \textbf{7} & \textbf{7} & 5 & 5 & 5 \\
    Hypercube & \#Best & \textbf{7} & 3 & 5 & 2 & 2 & 2 & 3 \\
    (7 instances) & \#Optimal & 0 & 0 & 0 & 0 & 0 & 0 & 0 \\
     & \#New & - & - & - & 0 & 0 & 0 & 0 \\ \hline
     & \#Solved & \textbf{40} & \textbf{40} & \textbf{40} & \textbf{40} & \textbf{40} & \textbf{40} & \textbf{40} \\
    Caterpillar & \#Best & 2 & 4 & 20 & \textbf{26} & 11 & 13 & 20 \\
    (40 instances) & \#Optimal & 0 & 0 & 0 & \textbf{16} & 11 & 13 & \textbf{16} \\
     & \#New & - & - & - & 4 & 0 & 0 & 0 \\ \hline
     & \#Solved & \textbf{24} & \textbf{24} & \textbf{24} & \textbf{24} & \textbf{24} & \textbf{24} & \textbf{24} \\
    CBT & \#Best & 0 & 0 & \textbf{12} & 0 & \textbf{12} & 6 & 11 \\
    (24 instances) & \#Optimal & 0 & 0 & 0 & 0 & \textbf{12} & 6 & 11 \\
     & \#New& - & - & - & 0 & 0 & 0 & 0 \\ \hline
    \end{tabular} 
    }
\end{table*}

In general, Table~\ref{tab:summary-structured-graph-results} highlights the key difference between metaheuristic approaches and SAT-CAB, as well as CP and MIP-based approaches. Specifically, while metaheuristic approaches are effective at finding good CAB values, they do not provide a way to prove that these values are optimal. In contrast, SAT-CAB, as well as CP and MIP-based approaches, can demonstrate that no better CAB value exists, thereby proving the optimality of the found solution.

As presented in Table~\ref{tab:summary-structured-graph-results}, although SAT-CAB is not a strong approach for finding the best CAB values, it nevertheless demonstrates a strong capability in identifying and certifying optimal solutions. On the \textit{3D mesh} dataset, SAT-CAB proves the optimality of 3 instances, outperforming CPLEX\textsubscript{CP}, CPLEX\textsubscript{MIP}, and Gurobi, each of which proves optimality for only 2 instances. Moreover, SAT-CAB is the only approach that discovers new CAB values on this dataset, identifying 2 previously unknown values, whereas none of the CP or MIP-based solvers achieve this. On the \textit{caterpillar} dataset, SAT-CAB finds the best CAB values for 20 instances, which is slightly fewer than the 26 instances achieved by CPLEX\textsubscript{CP}. However, both approaches prove optimality for the same number of instances, indicating that SAT-CAB remains competitive in terms of certification power. A similar trend can be observed on the \textit{CBT} dataset, where SAT-CAB proves optimality for 11 instances, closely matching the 12 optimal solutions obtained by CPLEX\textsubscript{MIP} and substantially outperforming the others.

\subsection{Evaluation Summary}

The experimental results presented above demonstrate the effectiveness of the proposed SAT-CAB approach for solving the Cyclic Anti-bandwidth Problem. On structured graphs, SAT-CAB demonstrates performance that is competitive with previously studied metaheuristic approaches, including MACAB, HABC-CAB, and MS-GVNS, on medium-sized instances while being less effective on larger graphs. Moreover, for the instance $P_{6\times6\times8}$, SAT-CAB identifies a new CAB value that exceeds the previously conjectured value in MACAB \cite{bansal2011memetic} (see Appendix~\ref{secA1}). On \textit{random connected graph} instances, SAT-CAB also consistently outperforms the CP and MIP-based approaches provided by CPLEX and Gurobi in terms of finding the best CAB values.

Notably, on the real-world \textit{Harwell–Boeing graph} dataset, SAT-CAB outperforms not only metaheuristic methods such as MACAB, HABC-CAB, HACO-CAB, and MS-GVNS, but also exact solvers, including CPLEX and Gurobi. Within a time limit of 150 seconds, SAT-CAB solves 23 out of 24 instances, proving optimality for 14 instances and identifying new CAB values for 8 instances. When the time limit is extended to 1800 seconds, SAT-CAB proves optimality for two additional instances and discovers seven further new CAB values, increasing the totals to 16 proven optimal CAB values and 10 new CAB values, respectively.

On the other hand, the experimental results also reveal several limitations of SAT-CAB that warrant further investigation and analysis. The first limitation lies in the size of the memory required for encoding and solving, especially for large graphs such as \textit{3D mesh} or \textit{hypercube}. Improving both the encoding process and the SAT solving strategy may help enhance SAT-CAB, making it a more promising approach for handling very large instances.

In addition, the parallel search strategy employed in SAT-CAB remains relatively naive, as it relies solely on a binary-search-based mechanism to determine the search order of CAB values without considering the structural characteristics of the graph, such as the number of vertices, the number of edges, or the degree distribution. Consequently, SAT-CAB performs less effectively on certain instances with wide bounds, such as S-sherman4 and V-nos6 in the \textit{Harwell-Boeing graph} dataset when compared with metaheuristic approaches that leverage parameter tuning to optimize their search processes.

\section{Conclusion}\label{sec5}
This paper introduces SAT-CAB, the first exact approach for the Cyclic Antibandwidth Problem on general graphs and the first method to address CABP within a SAT solving framework. In contrast to existing state-of-the-art techniques, which are predominantly heuristic or metaheuristic, the proposed approach systematically explores the solution space and provides formal guaranties of global optimality.

A key contribution of this work is the observation that CABP can be naturally modeled through a set of At-Most-One (AMO) constraints over consecutive variables arranged in a cycle. Building on this insight, we proposed a novel and compact SAT encoding specifically tailored to the structure of CABP. Rather than encoding each AMO constraint independently, the proposed encoding decomposes them into interrelated blocks and encodes these blocks jointly as a unified set of constraints. This strategy significantly reduces the number of auxiliary variables and clauses compared to generic encodings, enabling modern SAT solvers to handle instances of practical relevance.

Extensive computational experiments demonstrate that SAT-CAB advances the current state of the art by tightening known bounds and establishing several new best-known bounds for CABP. More importantly, the proposed approach is able to prove global optimal cyclic antibandwidth values for a substantial set of benchmark instances, for which no previous method was capable of certifying optimality.
Despite these promising results, the scalability of SAT-CAB remains limited for large or dense graphs, as the size of the generated SAT formulas may become prohibitive due to memory constraints. Addressing this limitation through more compact encodings, symmetry-breaking techniques, heuristic guidance, or hybrid SAT–metaheuristic frameworks constitutes an important direction for future research.

Overall, this work helps bridge the gap between heuristic and exact approaches for CABP, highlights the potential of SAT solving as an effective exact framework for this problem, and provides a useful reference for future research on exact, hybrid, and SAT-based methods for cyclic graph labeling problems.

\backmatter

\bmhead{Author Contributions}
Hieu Truong Xuan: conceptualization, methodology, implementation, data curation, visualization, writing - original draft, review \& editing;\\
Khanh To Van: conceptualization, investigation, methodology, formal analysis, supervision, funding acquisition, writing - original draft, review \& editing.

\section*{Declarations}

\bmhead{Ethics approval} Not applicable.
\bmhead{Consent to participate} Not applicable. 
\bmhead{Consent for publication} Not applicable.  
\bmhead{Conflict of interest} Not applicable.

\begin{appendices}

\section{Detail of the experimental datasets}\label{secA1}

This appendix provides descriptions of the datasets used in our experiments, along with their bounds of CAB values.

\begin{itemize}
    \item \textbf{Three-dimensional mesh} (20 instances). \\
    A \textit{three-dimensional mesh} (\textit{3D mesh}) $P_{n_1 \times n_2 \times n_3}$ with $n$ vertices is defined as the Cartesian product of three paths, $P_{n_1} \times P_{n_2} \times P_{n_3}$, where $n = n_1 \cdot n_2 \cdot n_3$. The size of \textit{three-dimensional mesh}, measured by the number of vertices $n$, ranges from 12 to 3600. 
    
    \textbf{\textit{Bound of CAB value:}} The lower bound for the CAB value of the \textit{3D mesh} instances is fixed at 2, whereas the upper bound is chosen as the value immediately above the conjectured CAB value reported by Bansal et~al.~\cite{bansal2011memetic} (see Table~\ref{tab:conjectured_3D_mesh}).

    \begin{table*}[ht!]
    \centering
    \begin{minipage}{0.7\textwidth}
    \caption{Conjectured CAB value of 3D mesh instances. Source: Bansal et al.~\cite{bansal2011memetic}.}
    \label{tab:conjectured_3D_mesh}
    \centering
    \begin{tabular}{cccc}
    \hline
    $n_1$ & $n_2$ & $n_3$ & Conjectured CAB value of $P_{n_1\times n_2\times n_3}$ \rule{0pt}{12pt}  \\ \hline
    2 & 2 & $3$--$500$ & $2(n_3 - 1)$ \rule{0pt}{12pt} \\
    3 & 3 & $3$--$400$ & 
        $\begin{cases}
          \dfrac{9n_3 - 8}{2}, & \text{if $n_3$ is even} \\
          \dfrac{9(n_3 - 1)}{2}, & \text{if $n_3$ is odd}
        \end{cases}$ \rule{0pt}{24pt} \\ 
    4 & 4 & $5$--$200$ & $8(n_3 - 1)$ \rule{0pt}{12pt} \\
    5 & 5 & $7$--$100$ & 
        $\begin{cases}
          \dfrac{25n_3 - 26}{2}, & \text{if $n_3$ is even} \\
          \dfrac{25n_3 - 27}{2}, & \text{if $n_3$ is odd}
        \end{cases}$ \rule{0pt}{24pt} \\
    6 & 6 & $8$--$100$ & $18n_3 - 19$ \rule{0pt}{12pt} \\ \hline
    \end{tabular}
    \end{minipage}
    \end{table*}

    \item \textbf{Double stars} (20 instances). \\
    A \textit{double stars} $s(n_1, n_2)$ is formed by combining two stars: the first with $n_1$ vertices and the second with $n_2$ vertices. In each star, a central vertex is connected to all other vertices. Additionally, the two central vertices of two stars are connected to each other. The size of \textit{double stars} ranges from 20 to 180 vertices.

    \textbf{\textit{Bound of CAB value:}} Similar to the \textit{3D mesh} instances, the lower bound for the CAB value of the \textit{double stars} instances is fixed at 2, whereas the upper bound is set to the value immediately above the conjectured CAB value proposed by Bansal et~al.~\cite{bansal2011memetic}, which is calculated using formula~\eqref{eq:ub_double_star}.

    \begin{equation}
        cab(s(n_1, n_2)) = 
        \begin{cases}
            \lfloor \frac{n_1}{2} \rfloor & \text{, if } n_1 = n_2 \rule{0pt}{12pt}\\
            \lceil \frac{min(n_1, n_2)}{2} \rceil & \text{, otherwise} \rule{0pt}{12pt}
        \end{cases}
        \label{eq:ub_double_star}
    \end{equation}
    
    \item \textbf{Hypercube} (7 instances). \\
    A \textit{hypercube} of dimension $d$, denoted by $Q_d$, is a $d$-dimensional graph with $2^{d}$ vertices, where each vertex is connected to exactly $d$ others. Consequently, $Q_d$ contains $d \cdot 2^{\,d-1}$ edges. In our study, we consider seven \textit{hypercube} instances with dimensions ranging from 4 to 10.

    \textbf{\textit{Bound of CAB value:}} The upper and lower bounds of the CAB value for the \textit{hypercube} instances are determined based on the corresponding Anti-bandwidth values, as given by Formula \eqref{eq:ab_to_cab_value} \cite{miller1989separation, raspaud2009antibandwidth}.  
    In addition, the formula for computing the Anti-bandwidth value for \textit{hypercube} instances was proposed by Wang et al. \cite{wang2009explicit}, as presented in Formula \eqref{eq:ab_value_of_hypercubes}.

    \begin{equation}
        \frac{1}{2}ab(Q_d) \leq cab(Q_d) \leq ab(Q_d)
        \label{eq:ab_to_cab_value}
    \end{equation}

    \begin{equation}
        ab(Q_d) = 2 ^ {d-1} - \sum_{m = 0} ^ {d - 2} \binom{m}{\lfloor m/2 \rfloor}.
        \label{eq:ab_value_of_hypercubes}
    \end{equation}

    \item \textbf{Caterpillar} (40 instances). \\ 
    A \textit{caterpillar} $P_{n_1, n_2}$ consists of one backbone path $P_{n_1}$ with $n_1$ vertices and $n_1$ branch paths $P_{n_2}$ connected to each of $P_{n_1}$'s vertices. We consider 40 \textit{caterpillar} instances with sizes ranging from 20 to 1000 vertices.

    \textbf{\textit{Bound of CAB value:}} To the best of our knowledge, there is currently no existing work proposing analytical bounds for \textit{caterpillar} instances. Therefore, the lower and upper bounds of the CAB value for \textit{caterpillar} instances are set to 2 and $\frac{n}{2}$, respectively.
    
    \item \textbf{Complete binary tree} (24 instances). \\
    A \textit{complete binary tree} $CBT_n$ of $n$ nodes is a rooted tree where all levels, except possibly the last one, are completely filled, and the nodes on the last level are aligned as far left as possible. In our experiments, we evaluated 24 instances of \textit{complete binary tree} with sizes ranging from 30 to 950 nodes..

    \textbf{\textit{Bound of CAB value:}} Similarly to the \textit{caterpillar} instances, the \textit{complete binary tree} instances are also evaluated with lower and upper bounds of 2 and $\frac{n}{2}$, respectively.

    \item \textbf{Harwell-Boeing graph} (24 instances). \\
    This dataset consists of 24 instances derived from the Harwell-Boeing Sparse Matrix Collection \cite{hbdataset}, which arises from real-world scientific and engineering applications such as structural mechanics, circuit simulation, and optimization. Among these 24 instances, there are 12 small graphs with sizes ranging from 30 to 118 vertices and 46 to 281 edges, while the other 12 are larger graphs with 420 to 715 vertices and 586 to 3720 edges.
    
    \textbf{\textit{Bound of CAB value:}} The bounds of the CAB value for the \textit{Harwell–Boeing graph} instances are determined based on the bounds of the Antibandwidth (AB) value, similar to the \textit{hypercube} instances (see Formula \ref{eq:ab_to_cab_value}). Specifically, the lower bound of the CAB value is taken as half of the lower bound of the AB value, while the upper bound is set equal to the upper bound of the AB value. The AB bounds for the \textit{Harwell–Boeing graph} instances are adopted from the work of Fazekas et al. \cite{fazekas2020duplex}.  
    Details for each instance in the \textit{Harwell–Boeing graph} dataset are presented in Table \ref{tab:harwell_boeing_problems_detail}, including the graph names, number of vertices $|V|$, number of edges $|E|$, lower and upper bounds of AB value ($AB_{lb}$ and $AB_{ub}$), and lower and upper bounds of CAB value ($CAB_{lb}$ and $CAB_{ub}$).

    \begin{table*}[ht!]
        \caption{Harwell Boeing Sparse Matrix graphs.}
        \label{tab:harwell_boeing_problems_detail}
        \centering
        \begin{minipage}[t]{0.48\textwidth}
        \resizebox{\textwidth}{!}{
        \begin{tabular}{lcccccc}
        \hline
        Instance & $\left|V\right|$ & $\left|E\right|$ & $AB_{lb}$ & $AB_{ub}$ & $CAB_{lb}$ & $CAB_{ub}$ \rule{0pt}{10pt} \\ \hline \hline
        A-pores\_1 & 30 & 103 & 6 & 8 & 3 & 8 \rule{0pt}{10pt} \\ \hline
        B-ibm32 & 32 & 90 & 9 & 9 & 5 & 9 \rule{0pt}{10pt} \\ \hline
        C-bcspwr01 & 39 & 46 & 16 & 17 & 8 & 17 \rule{0pt}{10pt} \\ \hline
        D-bcsstk01 & 48 & 176 & 8 & 9 & 4 & 9 \rule{0pt}{10pt} \\ \hline
        E-bcspwr02 & 49 & 59 & 21 & 22 & 11 & 22 \rule{0pt}{10pt} \\ \hline
        F-curtis54 & 54 & 124 & 12 & 13 & 6 & 13 \rule{0pt}{10pt} \\ \hline
        G-will57 & 57 & 127 & 12 & 14 & 6 & 14 \rule{0pt}{10pt} \\ \hline
        H-impcol\_b & 59 & 281 & 8 & 8 & 4 & 8 \rule{0pt}{10pt} \\ \hline
        I-ash85 & 85 & 219 & 19 & 27 & 10 & 27 \rule{0pt}{10pt} \\ \hline
        J-nos4 & 100 & 247 & 32 & 40 & 16 & 40 \rule{0pt}{10pt} \\ \hline
        K-dwt\_\_234 & 117 & 162 & 46 & 58 & 23 & 58 \rule{0pt}{10pt} \\ \hline
        L-bcspwr03 & 118 & 179 & 39 & 39 & 20 & 39 \rule{0pt}{10pt} \\ \hline
        \end{tabular}
        }
        \end{minipage}
        \hfill
        \begin{minipage}[t]{0.49\textwidth}
        \resizebox{\textwidth}{!}{
        \begin{tabular}{lcccccc}
        \hline
        Instance & $\left|V\right|$ & $\left|E\right|$ & $AB_{lb}$ & $AB_{ub}$ & $CAB_{lb}$ & $CAB_{ub}$ \rule{0pt}{10pt} \\ \hline \hline
        M-bcsstk06 & 420 & 3720 & 28 & 72 & 14 & 72 \rule{0pt}{10pt} \\ \hline
        N-bcsstk07 & 420 & 3720 & 28 & 72 & 14 & 72 \rule{0pt}{10pt} \\ \hline
        O-impcol\_d & 425 & 1267 & 91 & 173 & 46 & 173 \rule{0pt}{10pt} \\ \hline
        P-can\_\_445 & 445 & 1682 & 78 & 120 & 39 & 120 \rule{0pt}{10pt} \\ \hline
        Q-494\_bus & 494 & 586 & 219 & 246 & 110 & 246 \rule{0pt}{10pt} \\ \hline
        R-dwt\_\_503 & 503 & 2762 & 46 & 71 & 23 & 71 \rule{0pt}{10pt} \\ \hline
        S-sherman4 & 546 & 1341 & 256 & 272 & 128 & 272 \rule{0pt}{10pt} \\ \hline
        T-dwt\_\_592 & 592 & 2256 & 103 & 150 & 52 & 150 \rule{0pt}{10pt} \\ \hline
        U-662\_bus & 662 & 906 & 219 & 220 & 110 & 220 \rule{0pt}{10pt} \\ \hline
        V-nos6 & 675 & 1290 & 326 & 337 & 163 & 337 \rule{0pt}{10pt} \\ \hline
        W-685\_bus & 685 & 1282 & 136 & 136 & 68 & 136 \rule{0pt}{10pt} \\ \hline
        X-can\_\_715 & 715 & 2975 & 112 & 142 & 56 & 142 \rule{0pt}{10pt} \\ \hline
        \end{tabular}
        }
        \end{minipage}
    \end{table*}

    \item \textbf{Random connected graph} (40 instances). \\
    These graphs are generated by randomly adding edges between vertex pairs in such a way that the resulting graph is connected. This dataset comprises 40 instances, with graph sizes ranging from 100 to 200 vertices and 200 to 2000 edges. In contrast to structured graph families like \textit{3D mesh}, \textit{double stars}, or \textit{hypercube}, these graphs display irregular topologies without any inherent geometric or hierarchical patterns.

    \textbf{\textit{Bound of CAB value:}} The lower and upper bounds of the CAB value for \textit{random connected graph} instances are set to 2 and $\frac{n}{2}$, respectively.
    
\end{itemize}

\section{Experimental results of structured graph datasets}
\label{sec:appendB}

This appendix reports the detailed experimental results for each instance on the structured datasets, which consist of \textit{3D mesh}, \textit{double stars}, \textit{hypercube}, \textit{caterpillar}, and \textit{complete binary tree}. In Tables~\ref{tab:result_3dmesh_150s}, \ref{tab:result_hypercubes_150s}, and \ref{tab:result_double_stars_150s}, the column Conjectured Value refers to the conjectured CAB value of each instance, as reported by Bansal et al.~\cite{bansal2011memetic}. Note that since this value is conjectured, it may be less than the actual optimal CAB value. 

\begin{table*}[ht!]
    \caption{Experimental results of 3D mesh instances.} 
    \label{tab:result_3dmesh_150s}  
    \centering
    \resizebox{\textwidth}{!}{
        \begin{tabular}{lc|ccccccc}
        \hline
        \multicolumn{1}{c}{\multirow{2}{*}{\textbf{Instance}}} & \multicolumn{1}{l|}{\textbf{Conjectured}} & \multicolumn{7}{c}{\textbf{TO = 150s}} \\ \cline{3-9} 
        \multicolumn{1}{c}{} & \textbf{Value} & \textbf{MACAB} & \textbf{HABC-CAB} & \textbf{MS-GVNS} & \textbf{CPLEX\textsubscript{CP}} & \textbf{CPLEX\textsubscript{MIP}} & \textbf{Gurobi} & \textbf{SAT-CAB} \\ \hline
        $P_{2\times2\times3}$ & 4 & 4 & 4 & 4 & \textbf{4*} & \textbf{4*} & \textbf{4*} & \textbf{4*} \\
        $P_{2\times2\times168}$ & 334 & 333 & 332 & 332 & 239 & 6 & 5 & \textbf{334*} \\
        $P_{2\times2\times335}$ & 668 & \textbf{667} & 665 & 665 & 436 & TO & 3 & MO \\
        $P_{2\times2\times500}$ & 998 & \textbf{997} & 995 & 995 & 93 & TO & TO & MO \\
        $P_{3\times3\times3}$ & 9 & 9 & 9 & 9 & \textbf{9*} & \textbf{9*} & \textbf{9*} & \textbf{9*} \\
        $P_{3\times3\times135}$ & 603 & \textbf{602} & 599 & 597 & 404 & TO & 3 & 38 \\
        $P_{3\times3\times270}$ & 1211 & \textbf{1209} & 1206 & 1203 & 47 & TO & TO & MO \\
        $P_{3\times3\times400}$ & 1781 & \textbf{1779} & 1775 & 1755 & 14 & TO & TO & MO \\
        $P_{4\times4\times5}$ & 32 & \textbf{32} & \textbf{32} & \textbf{32} & \textbf{32} & 30 & 26 & \textbf{32} \\
        $P_{4\times4\times68}$ & 536 & \textbf{535} & 532 & 525 & 236 & TO & 3 & MO \\
        $P_{4\times4\times137}$ & 1088 & \textbf{1087} & 1078 & 1072 & 72 & TO & TO & MO \\
        $P_{4\times4\times200}$ & 1592 & \textbf{1588} & 1586 & 1503 & 32 & TO & TO & MO \\
        $P_{5\times5\times7}$ & 74 & 74 & 74 & \textbf{75} & \textbf{75} & 23 & 23 & 70 \\
        $P_{5\times5\times35}$ & 424 & \textbf{424} & 422 & 407 & 292 & 4 & 3 & 27 \\
        $P_{5\times5\times70}$ & 862 & \textbf{861} & 854 & 840 & 116 & TO & TO & MO \\
        $P_{5\times5\times100}$ & 1237 & \textbf{1236} & 1233 & 1214 & 54 & TO & TO & MO \\
        $P_{6\times6\times8}$ & 125 & 125 & 125 & 124 & 121 & 14 & 12 & \textbf{126} \\
        $P_{6\times6\times36}$ & 629 & \textbf{629} & 625 & 603 & 204 & TO & TO & MO \\
        $P_{6\times6\times72}$ & 1277 & \textbf{1276} & 1272 & 1254 & 47 & TO & TO & MO \\
        $P_{6\times6\times100}$ & 1796 & \textbf{1794} & 1788 & 1671 & 18 & TO & TO & MO \\ \hline
        \multicolumn{2}{l|}{\textbf{\#Solved}} & \textbf{20} & \textbf{20} & \textbf{20} & \textbf{20} & 7 & 10 & 8 \\
        \multicolumn{2}{l|}{\textbf{\#Best}} & \textbf{15} & 1 & 2 & 4 & 2 & 2 & 5 \\
        \multicolumn{2}{l|}{\textbf{\#Optimal}} & 0 & 0 & 0 & 2 & 2 & 2 & \textbf{3} \\ 
        \multicolumn{2}{l|}{\textbf{\#New}} & - & - & - & 0 & 0 & 0 & \textbf{2} \\ \hline
        \end{tabular}
    }
\end{table*}

\begin{table*}[ht!]
    \caption{Experimental results of hypercube instances.} 
    \label{tab:result_hypercubes_150s}  
    \centering
    \resizebox{\textwidth}{!}{
        \begin{tabular}{lc|ccccccc}
        \hline
        \multicolumn{1}{c}{\multirow{2}{*}{\textbf{}}} & \textbf{Conjectured} & \multicolumn{7}{c}{\textbf{TO = 150s}} \\ \cline{3-9} 
        \multicolumn{1}{c}{} & \textbf{Value} & \textbf{MACAB} & \textbf{HABC-CAB} & \textbf{MS-GVNS} & \textbf{CPLEX\textsubscript{CP}} & \textbf{CPLEX\textsubscript{MIP}} & \textbf{Gurobi} & \textbf{SAT-CAB} \\ \hline
        $Q_{4}$ & 4 & \textbf{4} & \textbf{4} & \textbf{4} & \textbf{4} & \textbf{4} & \textbf{4} & \textbf{4} \\
        $Q_{5}$ & 9 & \textbf{9} & \textbf{9} & \textbf{9} & \textbf{9} & \textbf{9} & \textbf{9} & \textbf{9} \\
        $Q_{6}$ & 19 & \textbf{19} & \textbf{19} & \textbf{19} & 18 & 17 & 17 & \textbf{19} \\
        $Q_{7}$ & 41 & \textbf{41} & 40 & \textbf{41} & 38 & 33 & 30 & 36 \\
        $Q_{8}$ & 85 & \textbf{85} & 83 & \textbf{85} & 59 & 45 & 46 & 51 \\
        $Q_{9}$ & 178 & \textbf{177} & 167 & 173 & 94 & TO & TO & TO \\
        $Q_{10}$ & 364 & \textbf{363} & 319 & 351 & 138 & TO & TO & MO \\ \hline
        \multicolumn{2}{l|}{\textbf{\#Solved}} & \textbf{7} & \textbf{7} & \textbf{7} & \textbf{7} & 5 & 5 & 5 \\
        \multicolumn{2}{l|}{\textbf{\#Best}} & \textbf{7} & 3 & 5 & 2 & 2 & 2 & 3 \\
        \multicolumn{2}{l|}{\textbf{\#Optimal}} & 0 & 0 & 0 & 0 & 0 & 0 & 0 \\
        \multicolumn{2}{l|}{\textbf{\#New}} & - & - & - & 0 & 0 & 0 & 0 \\ \hline
        \end{tabular}
    }
\end{table*}

\begin{table*}[ht!]
    \caption{Experimental results of double stars instances.} 
    \label{tab:result_double_stars_150s}  
    \centering
    \resizebox{\textwidth}{!}{
        \begin{tabular}{lc|ccccccc}
        \hline
        \multicolumn{1}{c}{\multirow{2}{*}{\textbf{Instance}}} & \textbf{Conjectured} & \multicolumn{7}{c}{\textbf{TO = 150s}} \\ \cline{3-9} 
        \multicolumn{1}{c}{} & \textbf{Value} & \textbf{MACAB} & \textbf{HABC-CAB} & \textbf{MS-GVNS} & \textbf{CPLEX\textsubscript{CP}} & \textbf{CPLEX\textsubscript{MIP}} & \textbf{Gurobi} & \textbf{SAT-CAB} \\ \hline
        $s_{15,5}$ & 3 & 3 & 2 & 3 & 3 & \textbf{3*} & \textbf{3*} & \textbf{3*} \\
        $s_{15,7}$ & 4 & 4 & 3 & 4 & 4 & \textbf{4*} & \textbf{4*} & \textbf{4*} \\
        $s_{15,10}$ & 5 & 5 & 5 & 5 & 5 & \textbf{5*} & \textbf{5*} & \textbf{5*} \\
        $s_{15,12}$ & 6 & 6 & 6 & 6 & 6 & \textbf{6*} & \textbf{6*} & 6 \\
        $s_{30,20}$ & 10 & 10 & 10 & 10 & 10 & \textbf{10*} & \textbf{10*} & 10 \\
        $s_{30,25}$ & 13 & 13 & 12 & 13 & 13 & \textbf{13*} & \textbf{13*} & 13 \\
        $s_{35,20}$ & 10 & 10 & 10 & 10 & 10 & \textbf{10*} & \textbf{10*} & 10 \\
        $s_{35,25}$ & 13 & 13 & 12 & 13 & 13 & \textbf{13*} & \textbf{13*} & 13 \\
        $s_{40,20}$ & 10 & 10 & 10 & 10 & 10 & \textbf{10*} & \textbf{10*} & 10 \\
        $s_{40,25}$ & 13 & 13 & 12 & 13 & 13 & \textbf{13*} & \textbf{13*} & 13 \\
        $s_{40,30}$ & 15 & 15 & 15 & 15 & 15 & \textbf{15*} & \textbf{15*} & 15 \\
        $s_{50,20}$ & 10 & 10 & 10 & 10 & 10 & \textbf{10*} & \textbf{10*} & 10 \\
        $s_{50,25}$ & 13 & 13 & 12 & 13 & 13 & \textbf{13*} & \textbf{13*} & 13 \\
        $s_{50,30}$ & 15 & 15 & 15 & 15 & 15 & \textbf{15*} & \textbf{15*} & 15 \\
        $s_{100,20}$ & 10 & 10 & 10 & 10 & 10 & \textbf{10*} & \textbf{10*} & 10 \\
        $s_{100,25}$ & 13 & 13 & 12 & 13 & 13 & \textbf{13*} & \textbf{13*} & 13 \\
        $s_{100,30}$ & 15 & 15 & 15 & 15 & 15 & \textbf{15*} & \textbf{15*} & 15 \\
        $s_{150,20}$ & 10 & 10 & 10 & 10 & 10 & \textbf{10*} & \textbf{10*} & 10 \\
        $s_{150,25}$ & 13 & 13 & 12 & 13 & 13 & \textbf{13*} & \textbf{13*} & 13 \\
        $s_{150,30}$ & 15 & 15 & 15 & 15 & 15 & \textbf{15*} & \textbf{15*} & 15 \\ \hline
        \multicolumn{2}{l|}{\textbf{\#Solved}} & \textbf{20} & \textbf{20} & \textbf{20} & \textbf{20} & \textbf{20} & \textbf{20} & \textbf{20} \\
        \multicolumn{2}{l|}{\textbf{\#Best}} & 0 & 0 & 0 & 0 & \textbf{20} & \textbf{20} & 3 \\
        \multicolumn{2}{l|}{\textbf{\#Optimal}} & 0 & 0 & 0 & 0 & \textbf{20} & \textbf{20} & 3 \\ 
        \multicolumn{2}{l|}{\textbf{\#New}} & - & - & - & 0 & 0 & 0 & 0 \\ \hline
        \end{tabular}
    }
\end{table*}

\begin{table*}[ht!]
    \caption{Experimental results of complete binary tree instances.} 
    \label{tab:result_cbt_150s}  
    \centering
    \resizebox{\textwidth}{!}{
        \begin{tabular}{l|ccccccc}
        \hline
        \multicolumn{1}{c|}{\multirow{2}{*}{\textbf{Instance}}} & \multicolumn{7}{c}{\textbf{TO = 150s}} \\ \cline{2-8} 
        \multicolumn{1}{c|}{} & \textbf{MACAB} & \textbf{HABC-CAB} & \textbf{MS-GVNS} & \textbf{CPLEX\textsubscript{CP}} & \textbf{CPLEX\textsubscript{MIP}} & \textbf{Gurobi} & \textbf{SAT-CAB} \\ \hline
        $CBT_{30}$ & 10 & 12 & 12 & 12 & \textbf{12*} & \textbf{12*} & \textbf{12*} \\
        $CBT_{31}$ & 10 & 12 & 12 & 12 & \textbf{12*} & \textbf{12*} & \textbf{12*} \\
        $CBT_{32}$ & 11 & 13 & 13 & 13 & \textbf{13*} & \textbf{13*} & \textbf{13*} \\
        $CBT_{33}$ & 12 & 13 & 13 & 13 & \textbf{13*} & \textbf{13*} & \textbf{13*} \\
        $CBT_{34}$ & 13 & 14 & 14 & 14 & \textbf{14*} & \textbf{14*} & \textbf{14*} \\
        $CBT_{35}$ & 13 & 14 & 14 & 14 & \textbf{14*} & \textbf{14*} & \textbf{14*} \\
        $CBT_{45}$ & 15 & 19 & 19 & 19 & \textbf{19*} & 19 & \textbf{19*} \\
        $CBT_{46}$ & 15 & 19 & 19 & 19 & \textbf{19*} & 19 & \textbf{19*} \\
        $CBT_{47}$ & 15 & 20 & 20 & 20 & \textbf{20*} & 20 & \textbf{20*} \\
        $CBT_{48}$ & 16 & 20 & 20 & 20 & \textbf{20*} & 20 & \textbf{20*} \\
        $CBT_{49}$ & 17 & 21 & 21 & 21 & \textbf{21*} & 21 & \textbf{21*} \\
        $CBT_{50}$ & 18 & 21 & 21 & 21 & \textbf{21*} & 21 & 21 \\
        $CBT_{500}$ & 170 & 198 & \textbf{221} & 207 & 17 & 13 & 188 \\
        $CBT_{510}$ & 170 & 204 & \textbf{225} & 205 & 16 & 13 & 192 \\
        $CBT_{550}$ & 209 & 234 & \textbf{247} & 225 & 14 & 12 & 189 \\
        $CBT_{600}$ & 213 & 265 & \textbf{276} & 249 & 13 & 11 & 138 \\
        $CBT_{620}$ & 214 & 279 & \textbf{286} & 251 & 13 & 10 & 142 \\
        $CBT_{630}$ & 213 & 289 & \textbf{292} & 240 & 12 & 10 & 165 \\
        $CBT_{640}$ & 214 & 297 & \textbf{298} & 249 & 12 & 10 & 181 \\
        $CBT_{730}$ & 236 & 337 & \textbf{339} & 271 & 10 & 8 & 205 \\
        $CBT_{790}$ & 253 & 343 & \textbf{363} & 313 & 9 & 8 & 226 \\
        $CBT_{880}$ & 293 & 363 & \textbf{401} & 303 & 8 & 7 & 271 \\
        $CBT_{910}$ & 308 & 370 & \textbf{412} & 312 & 8 & 7 & 286 \\
        $CBT_{950}$ & 328 & 369 & \textbf{429} & 316 & 8 & 6 & 298 \\ \hline
        \textbf{\#Solved} & \textbf{24} & \textbf{24} & \textbf{24} & \textbf{24} & \textbf{24} & \textbf{24} & \textbf{24} \\
        \textbf{\#Best} & 0 & 0 & \textbf{12} & 0 & \textbf{12} & 6 & 11 \\
        \textbf{\#Optimal} & 0 & 0 & 0 & 0 & \textbf{12} & 6 & 11 \\
        \textbf{\#New} & - & - & - & 0 & 0 & 0 & 0 \\ \hline
        \end{tabular}
    }
\end{table*}

\begin{table*}[ht!]
    \caption{Experimental results of caterpillar instances.} 
    \label{tab:result_caterpillar_150s}  
    \centering
    \resizebox{\textwidth}{!}{
        \begin{tabular}{l|ccccccc}
        \hline
        \multicolumn{1}{c|}{\multirow{2}{*}{\textbf{Instance}}} & \multicolumn{7}{c}{\textbf{TO = 150s}} \\ \cline{2-8} 
        \multicolumn{1}{c|}{} & \textbf{MACAB} & \textbf{HABC-CAB} & \textbf{MS-GVNS} & \textbf{CPLEX\textsubscript{CP}} & \textbf{CPLEX\textsubscript{MIP}} & \textbf{Gurobi} & \textbf{SAT-CAB} \\ \hline
        $P_{5,4}$ & 8 & 8 & 8 & \textbf{8*} & \textbf{8*} & \textbf{8*} & \textbf{8*} \\
        $P_{5,5}$ & 10 & 11 & 11 & \textbf{11*} & \textbf{11*} & \textbf{11*} & \textbf{11*} \\
        $P_{5,6}$ & 13 & 13 & 13 & \textbf{13*} & \textbf{13*} & \textbf{13*} & \textbf{13*} \\
        $P_{5,7}$ & 15 & 16 & 16 & \textbf{16*} & \textbf{16*} & \textbf{16*} & \textbf{16*} \\
        $P_{9,4}$ & 15 & 16 & 16 & \textbf{16*} & \textbf{16*} & \textbf{16*} & \textbf{16*} \\
        $P_{9,5}$ & 19 & 20 & 20 & \textbf{20*} & \textbf{20*} & \textbf{20*} & \textbf{20*} \\
        $P_{9,6}$ & 23 & 24 & 25 & \textbf{25*} & \textbf{25*} & \textbf{25*} & \textbf{25*} \\
        $P_{9,7}$ & 28 & 29 & 29 & \textbf{29*} & 29 & \textbf{29*} & \textbf{29*} \\
        $P_{10,4}$ & 17 & 18 & 18 & \textbf{18*} & \textbf{18*} & \textbf{18*} & \textbf{18*} \\
        $P_{10,5}$ & 22 & 23 & 23 & \textbf{23*} & \textbf{23*} & \textbf{23*} & \textbf{23*} \\
        $P_{10,6}$ & 26 & 27 & 28 & \textbf{28*} & \textbf{28*} & \textbf{28*} & \textbf{28*} \\
        $P_{10,7}$ & 31 & 32 & 33 & \textbf{33*} & 32 & \textbf{33*} & \textbf{33*} \\
        $P_{15,4}$ & 27 & 28 & 28 & \textbf{28*} & \textbf{28*} & \textbf{28*} & \textbf{28*} \\
        $P_{15,5}$ & 34 & 35 & 35 & \textbf{35*} & 34 & 35 & \textbf{35*} \\
        $P_{15,6}$ & 41 & \textbf{42} & \textbf{42} & \textbf{42} & 40 & 40 & \textbf{42} \\
        $P_{15,7}$ & 48 & 49 & 50 & \textbf{50*} & 46 & 42 & \textbf{50*} \\
        $P_{20,4}$ & 36 & 38 & 38 & \textbf{38*} & 37 & 38 & \textbf{38*} \\
        $P_{20,5}$ & 46 & \textbf{47} & \textbf{47} & \textbf{47} & 44 & 41 & \textbf{47} \\
        $P_{20,6}$ & 56 & \textbf{57} & \textbf{57} & \textbf{57} & 52 & 48 & \textbf{57} \\
        $P_{20,7}$ & 66 & 66 & \textbf{67} & \textbf{67} & 50 & 47 & 66 \\
        $P_{20,10}$ & 94 & 95 & \textbf{96} & \textbf{96} & 42 & 37 & \textbf{96} \\
        $P_{20,15}$ & 142 & 142 & \textbf{145} & \textbf{145} & 29 & 24 & 144 \\
        $P_{20,20}$ & 190 & 190 & 191 & \textbf{195} & 21 & 17 & 157 \\
        $P_{20,25}$ & \textbf{240} & 238 & \textbf{240} & 193 & 17 & 13 & 188 \\
        $P_{25,10}$ & 118 & 119 & 120 & \textbf{121} & 35 & 29 & 120 \\
        $P_{25,15}$ & 179 & 179 & \textbf{182} & 178 & 22 & 18 & 176 \\
        $P_{25,20}$ & 239 & 239 & \textbf{242} & 194 & 17 & 13 & 188 \\
        $P_{25,25}$ & 300 & 300 & \textbf{301} & 244 & 13 & 10 & 166 \\
        $P_{30,10}$ & 144 & 144 & 145 & \textbf{146} & 29 & 24 & 141 \\
        $P_{30,15}$ & 217 & 216 & \textbf{218} & 183 & 19 & 15 & 183 \\
        $P_{30,20}$ & 289 & 288 & \textbf{293} & 241 & 13 & 11 & 226 \\
        $P_{30,25}$ & 362 & 361 & \textbf{364} & 285 & 10 & 8 & 40 \\
        $P_{35,10}$ & 168 & 169 & 169 & \textbf{170} & 25 & 20 & 165 \\
        $P_{35,15}$ & 253 & 253 & \textbf{256} & 204 & 16 & 12 & 197 \\
        $P_{35,20}$ & \textbf{338} & \textbf{338} & \textbf{338} & 260 & 11 & 9 & 88 \\
        $P_{35,25}$ & 424 & 422 & \textbf{425} & 301 & 8 & 7 & 75 \\
        $P_{40,10}$ & 193 & 194 & \textbf{195} & 167 & 21 & 17 & 176 \\
        $P_{40,15}$ & 291 & 290 & \textbf{293} & 227 & 13 & 11 & 188 \\
        $P_{40,20}$ & 388 & 387 & \textbf{391} & 283 & 9 & 8 & 101 \\
        $P_{40,25}$ & 486 & 484 & \textbf{488} & 329 & 7 & 6 & 63 \\ \hline
        \textbf{\#Solved} & \textbf{40} & \textbf{40} & \textbf{40} & \textbf{40} & \textbf{40} & \textbf{40} & \textbf{40} \\
        \textbf{\#Best} & 2 & 4 & 20 & \textbf{26} & 11 & 13 & 20 \\
        \textbf{\#Optimal} & 0 & 0 & 0 & \textbf{16} & 11 & 13 & \textbf{16} \\ 
        \textbf{\#New} & - & - & - & 4 & 0 & 0 & 0 \\ \hline
        \end{tabular}
    }
\end{table*}

\end{appendices}

\clearpage

\bibliography{sn-bibliography}


\end{document}